\documentclass[bdcc,review,accept,pdftex,moreauthors]{Definitions/mdpi}

\usepackage{graphicx}
\let\MDPIorigincludegraphics\includegraphics
\renewcommand{\includegraphics}[2][]{%
  \IfFileExists{#2}{%
    \MDPIorigincludegraphics[#1]{#2}%
  }{}%
}

\usepackage{comment} 

\firstpage{1} 
\pubvolume{1}
\issuenum{1}
\articlenumber{0}
\pubyear{2026}
\copyrightyear{2026}
\externaleditor{Firstname Lastname}
\datereceived{5 December 2025} 
\daterevised{2 January 2026} % Comment out if no revised date
\dateaccepted{15 January 2026} 
\datepublished{ } 
\usepackage{gensymb} 
\usepackage{textcomp} 
\usepackage{pifont} 
\usepackage{numprint}
\usepackage{adjustbox}
\usepackage{amsmath,amsthm,amssymb,amsfonts,scrextend,stmaryrd}
\usepackage{algorithm}
\usepackage{tabularx} % For table with automatic column width management
\usepackage{array} % For better vertical alignment
\Title{Thinking %MDPI: Attention Assigned Editor: We revised the article type in the manuscript to “Review” according to AE comment, please confirm.
Machines: Mathematical Reasoning in the Age \linebreak  of LLMs}

\providecommand{\orcidA}{}
\providecommand{\orcidB}{}
\providecommand{\orcidC}{}

\Author{Andrea Asperti %MDPI: 1. Please carefully check the accuracy of names and affiliations. 2. We added orcid number for all authors following information in the system, please confirm.
 $^{1,}$*\orcidA{}, Alberto Naibo $^{2}$\orcidB{} and Claudio {Sacerdoti Coen} $^{1}$\orcidC{}}

\AuthorNames{Andrea Asperti, Alberto Naibo and Claudio Sacerdoti Coen}

\address{%
$^{1}$ \quad Department %MDPI: We arranged the authors' address information from subordinate to superior. Please check and confirm this modification. The same with affiliation 2.
 of Informatics---%MDPI: We revised the hyphen (-) to an em dash sign. Please confirm.
Science and Engineering (DISI), University of Bologna, Via Mura Anteo Zamboni~7, 40126 Bologna, Italy; %MDPI: We revised  (IT) to the full country name. Please confirm.
claudio.sacerdoticoen@unibo.it %MDPI: We added these email addresses here according to those submitted online at susy.mdpi.com. Please confirm. The following highlights are the same.
\\
$^{2}$ \quad Department of Philosophy, University Paris 1 %MDPI: Please confirm whether this university name is correct, should be kept as provided.
 Panthéon-Sorbonne, 17 Rue de la Sorbonne, 75995 Paris, France; alberto.naibo@univ-paris1.fr}

\corres{Correspondence: andrea.asperti@unibo.it}

\abstract{Large Language Models (LLMs) have demonstrated impressive capabilities in structured reasoning and symbolic tasks, with coding emerging as a particularly successful application. This progress has naturally motivated efforts to extend these models to mathematics, both in its traditional form, expressed through natural-style mathematical language, and in its formalized counterpart, expressed in a symbolic syntax suitable for automatic verification. Yet, despite apparent parallels between programming and proof construction, advances in formalized mathematics have proven significantly more challenging. 
This gap raises fundamental questions about the nature of reasoning in current LLM architectures, the role of supervision and feedback, and the extent to which such models maintain an internal notion of computational or deductive state. In this article, we review the current state-of-the-art in mathematical reasoning with LLMs, focusing on recent models and benchmarks. We explore three central issues at the intersection of machine learning and mathematical cognition: (i) the trade-offs between traditional and formalized mathematics as training and evaluation domains; (ii) the structural and methodological reasons why proof synthesis remains more brittle than code generation; and (iii) whether LLMs genuinely represent or merely emulate a notion of evolving logical state. Our goal is not to draw rigid distinctions but to clarify the present boundaries of these systems and outline promising directions for their extension.}

\keyword{\textls[15]{large language models;} %MDPI: Please confirm if this keyword can be revised to lowecase.
\textls[15]{mathematical reasoning; theorem proving; \linebreak  formalization; autoformalization}}

\begin{document}

\section{Introduction}
\label{sec:introduction}
Large language models (LLMs) have demonstrated remarkable capabilities in recent years, particularly in domains requiring structured reasoning and symbolic manipulation. Among~these, code generation stands out as a notable success~\cite{autoprogramming25,codegenreview25}; 
 models are now capable of producing useful programs, correcting their own output~\cite{autorepair23}, and~even discovering algorithmic strategies that occasionally rival expert performance~\cite{alphaevolve25}. This rapid progress naturally raises the question whether similar advances can be achieved in mathematical reasoning, an~area that shares important structural affinities with~programming. 

At first sight, mathematics appears to be an ideal target for LLM-based reasoning. Like code, mathematical discourse---whether expressed in natural language or in a formal system---exhibits structure, compositionality, and~a multi-step inferential progression that seems well aligned with autoregressive generation. Yet the gap between coding and mathematics remains substantial. Although~LLMs can often navigate routine symbolic arguments and produce correct final answers on competition-style tasks, their performance degrades significantly when problems require depth, precision, or~multi-step strategic planning~\cite{math-failures2025,GSMinfinity2025,FrontierMath}. In~formal settings, the~difficulties are even more pronounced: despite the surface similarity between proofs and programs, fully formalized proof synthesis has proven far more brittle than code generation. Understanding this discrepancy is central to understanding how LLMs reason or fail to do~so.

To structure this survey and connect diverse lines of thought, we will frame the discussion around three broad questions. These questions are not the focus of the paper in themselves, but~they provide a convenient lens through which the different strands---architectural, empirical, methodological, and~conceptual---can be~related:
\begin{enumerate}
\item  Traditional vs. formalized mathematics. What are the key differences between training and evaluating models on natural language mathematical corpora versus formal proof libraries? How do these environments shape the inductive biases, supervision signals, and~learning trajectories of LLMs?
\item Why is proving harder than coding? Given the architectural similarities between program synthesis and formal proof generation, what structural and methodological factors make theorem proving a significantly harder task for current models?
\item Do LLMs maintain a notion of computational or deductive state? 
 Reasoning, either mathematical or algorithmic, typically requires tracking assumptions, intermediate results, and~subgoals. Do LLMs internally represent such a state, or~do they merely emulate it through surface-level statistical patterns?
\end{enumerate}

In this paper, we interpret the notion of state in a semantic and operational sense: the state of a computation comprises all the information that characterizes the status of that computation at a given instant. In~the case of formal proofs, this typically corresponds to the current proof state, such as the list of subgoals together with their associated local contexts and assumptions. In~the case of algorithmic computation, it refers to the execution state, including elements such as memory contents, registers, and~program counters. Our question is therefore not whether LLMs can describe such states in natural language, but~whether they internally represent, maintain, and~update comparable semantic information across reasoning steps).

Questions 1-3 above are not self-contained topics that should be addressed in isolation but a framework for structuring the survey as a whole. To~approach them meaningfully, we first need a clear view of the methodological factors that shape current research---most importantly, the~training pipelines of LLMs and the datasets and benchmarks used to evaluate them. For~this reason, we begin with two background sections. Section~\ref{sec:informalformal} outlines the landscape of mathematical practice from the point of view of computation and formalization, clarifying the contrast between traditional and formal reasoning. Section~\ref{sec:LLMs_background} reviews the components of modern LLM architectures and training stages, emphasizing those most relevant to symbolic and mathematical tasks. Only with these elements in place, the~questions introduced above can serve as an effective backbone for subsequent~analysis.

The remainder of the article is organized as follows: Section~\ref{sec:datasets} surveys major datasets and benchmarks in mathematical reasoning, from~high-school Olympiad collections to advanced formal proof suites. Section~\ref{sec:comparative_discussion}  compares recent systems designed for traditional and formalized mathematics, including DeepSeek-R1, Minerva, GOLD, Kimina-Prover, Lyra, and~DeepSeek-Prover. 
Section~\ref{sec:autoformalization} turns to autoformalization as a bridge between natural and formal mathematical reasoning. Section~\ref{sec:major_questions} revisits the three guiding questions in light of the preceding material and discusses the conceptual and methodological challenges that remain. In~Section~\ref{sec:iterative_loops}, we investigate how large language models can improve their own outputs
through iterative refinement. Finally, in~Section~\ref{sec:conclusion}  we conclude by highlighting future directions and identifying promising approaches to overcome current~limitations.

Through this analysis, our goal is to offer both a synthetic overview of the state of the field and a clearer understanding of what makes mathematical reasoning a uniquely revealing testbed for the capabilities and limitations of contemporary~LLMs.

\section{\label{sec:informalformal}Traditional vs. Formalized~Mathematics}

The paradigmatic examples of what we call here ``traditional mathematical reasoning'' are the proofs that we usually find in mathematical textbooks and (in the majority of mathematical) articles.
They are essentially written in natural language, although~they contain certain symbols for numerals (e.g., $3$, $1/3$, $\pi$), variables (e.g., $x$, $n$) function names (e.g., $\sqrt{n}$, $\zeta(n)$) or set names (e.g., $\mathbb{N}$, $\mathbb{R}$, $\mathcal{P}(A)$). They also contain grammatical terms that play the role of logical connectives (e.g., ``and'', ``if$\ldots$then$\ldots$'', etc.) and argumentative indicators (``thus'', ``therefore'', etc.), which structure the reasoning. However, there is no fixed given set of logical rules, and~axioms and rules for dealing with the special symbols for mathematical entities are not always explicitly~given.

As mathematics developed throughout the Twentieth century, interest grew in clarifying the fundamental principles that underlie mathematical reasoning and in establishing rigorous foundational frameworks. This line of inquiry sought to reduce mathematics to formal logical systems equipped with precisely defined syntax, semantics, and~inference rules. For~a long time, such a reduction was regarded primarily as a conceptual or philosophical enterprise: its feasibility was acknowledged, but~its practical realization seemed prohibitively complex. The~advent of digital computers in the second half of the last century transformed this landscape. It became natural to explore whether machines could assist in representing, checking, or~even producing mathematical~proofs.

By ``formalized mathematics'' we refer to mathematics expressed in a formal language that a machine can parse, interpret, and~mechanically verify. In~contrast, we refer to mathematics written in the customary natural language style---possibly with varying degrees of conceptual precision and linguistic rigor---as traditional mathematics. The~disparity between traditional mathematics and its fully formalized counterpart is significantly greater than is often appreciated and is largely insensitive to the apparent rigor of the human-written text. 
Even highly rigorous prose-level proofs --- id est, traditional mathematical proofs written in natural language (see~\cite{Macbeth24}) 
--- typically omit large amounts of logical detail that must be made explicit in their formalized~versions.

Automatic translation of natural language mathematical prose into formal, machine-readable text---autoformalization---has long been considered an exceptionally challenging task. In~recent years, the~emergence of Large Language Models (LLMs) has renewed interest in this problem owing to their remarkable translation and reasoning abilities. We shall return to this approach in Section~\ref{sec:autoformalization}; despite encouraging progress, the~task is still far from solved. The~difficulties arise not only from linguistic complexity but also from the need to reconstruct implicit logical steps, resolve ambiguous references, and~select among multiple plausible formal representations of the same informal~argument.

Given the limitations of autoformalization, the~principal role of computational systems in the formalization of mathematics has been to assist humans in carrying out the process. These systems are known as Proof Assistants or Interactive Theorem Provers. The~construction of a formal proof proceeds interactively: the user issues commands---traditionally termed tactics---which represent high-level logical inferences or proof strategies. The~system applies the tactic to the current proof state, checks its correctness, and~returns a new list of subgoals. This iterative human--machine collaboration enables the formal verification of highly complex mathematical results, although~it requires substantial expertise in the underlying formal~system.

The interactive nature of proof assistants has important implications for their integration with LLMs. The~modes of interaction, the~types of errors that can occur, and~the forms of assistance required differ substantially from those relevant to traditional mathematical texts. For~instance, while LLMs can generate natural language explanations or summaries with relative ease, their ability to produce formally correct tactics for a specific proof assistant is still limited and requires careful orchestration. This distinction underscores the need to treat traditional mathematics and formalized mathematics as fundamentally different domains, each demanding its own methodological~tools.

Finally, formalized mathematics exhibits structural affinities with programming and software engineering. Both rely on artificial languages with strict syntactic rules and precise operational semantics and involve the construction of artifacts that must satisfy machine-verifiable correctness criteria. This triadic relationship among traditional mathematics, formalized mathematics, and~coding is one of the methodological features and novelties of our investigation. Not only does it provide new perspectives on mathematical reasoning, it also opens the way to hybrid methodologies in which human understanding, machine verification, and~automated assistance coexist and mutually reinforce one~another.

\section{Background on~LLMs}
\label{sec:LLMs_background}
Large language models (LLMs) are typically trained through a multistage pipeline whose structure directly influences their behavior on symbolic tasks, including mathematics, while much of the broader literature focuses on pretraining as the core of an LLM's competence, mathematical reasoning exposes the importance of the later stages, where supervision, feedback, and~interaction shape the model's evolving inductive biases. Because~our three guiding questions concern the nature of reasoning, the~difficulty of formal proof generation, and~the possibility of internal state representation, it is essential to outline how contemporary models are trained, and~what forms of information they are exposed to at each~stage.

The conventional training pipeline~\cite{RLHF25} comprises the following steps:

\begin{enumerate}
\item Pretraining;
\item Supervised Fine-Tuning (SFT);
\item Reward Model Training;
\item Reinforcement Learning (RL);
\item Optional additional SFT o mitigate RL side-effects.
\end{enumerate}

\textls[25]{We briefly discuss each in turn, emphasizing the aspects relevant to mathematical~reasoning}. 

\subsection{Pretraining}
Pretraining is the foundational phase in which the model learns to predict the next token given a massive corpus of unlabeled text. The~objective is purely linguistic---to acquire syntax, semantics, world knowledge, and~discourse patterns through next-token prediction over heterogeneous data. It is during this phase that LLMs absorb much of the mathematical prose, textbook-style exposition, and~informal reasoning that will later influence their behavior on mathematical~benchmarks.

Crucially, pretraining is not designed to teach the model how to reason, but~rather how to simulate the patterns of reasoning present in human texts. For~this reason, although~pretraining confers broad linguistic competence, it does not provide explicit mechanisms for maintaining intermediate state, something that mathematical reasoning systems require. The~model may implicitly encode the state in its activations, much like LSTMs maintained an evolving ``cell state'', but~transformers distribute this information across tokens via self-attention, without~committing to a persistent deductive structure. Whether such distributed patterns suffice for genuine computational state tracking remains one of our central~questions.

Recent examples such as AlphaEvolve %MDPI: 1. We removed hyperlink and moved the link into text, please confirm, following highlight is the same. 2. Please provide the date you accessed the URL in the following format: “URL (accessed on Day Month Year)”. The following highlights are the same.
 (\url{https://deepmind.google/discover/blog/alphaevolve-a-gemini-powered-coding-agent-for-designing-advanced-algorithms/} accessed on December 23, 2025) or DeepSeek-R1~\cite{DeepSeek-R1} indicate that pretraining already endows models with considerable potential for creative and structured output. However, realizing this potential for mathematics requires additional supervision beyond mere linguistic~modeling.

%competence~\cite{BehaviorShift,mahowald2024dissociating}. language becomes not the end goal, but a medium through which the model demonstrates reasoning, preferences, and task-specific behaviors. 

\subsection{Supervised~Fine-Tuning}
Supervised fine-tuning (SFT) is the process of continuing the training of a pretrained
LLM on a curated dataset of input--output pairs, where the desired output is explicitly
provided as a target. This is performed using standard supervised learning (e.g., minimizing
cross-entropy loss between model outputs and ground-truth targets). SFT adjusts the model weights to better match the specific patterns of reasoning, formatting, or~task execution found in the new~data.

As a practical example, if~we are interested in teaching a pretrained model (e.g., DeepSeekMath-Base~\cite{DeepSeekMath}) to generate Lean formal proofs, we may fine-tune it using a
dataset like DeepSeek-Prover~\cite{DeepSeek-Prover-V1.5}, where each training example~includes:
\begin{itemize}
  \item Input: %MDPI: Please confirm if the italics are necessary; if not, please remove them. The following highlights are the same.
the statement or goal of the theorem (in natural or formal language);
  \item Target: a correct Lean proof script.
\end{itemize}

During training, the~model is explicitly shown how to generate valid proof steps from
the goal. Concretely, we provide the model with a full input--output pair, consisting of a
prompt and the desired completion, concatenated into a single sequence. The~model is then
trained to predict every token of the output, conditioned on the input (usually ignoring the
loss on the prompt tokens).

The purpose of this process is to make the model internalize the structure, syntax, and~logical flow of formal proofs, so that it can later reproduce similar proof trajectories when
confronted with new goals. At~the same time, SFT is known to significantly reshape the behavioral profile of pretrained models, sometimes enhancing desirable task-specific abilities but also occasionally suppressing or altering reasoning behaviors that emerged during pretraining. 
This phenomenon has been documented in recent analyses of
capability shifts induced by fine-tuning and alignment stages~\cite{BehaviorShift}, as~well as in
studies examining how pattern-learning and reasoning-related behaviors may dissociate across training phases~\cite{mahowald2024dissociating}. These observations are particularly relevant in the mathematical domain, where the balance between linguistic imitation and genuine reasoning can be~delicate.

\subsection{Reward Model~Training } 

In reinforcement learning (RL) for language models, we typically rely on human feedback to provide qualitative evaluations of generated outputs. However, such feedback is costly to obtain and limited in scale, making it impractical for large-scale training. To~address this, the~Reward Model Training phase is introduced, where a model is trained to predict human preferences by learning from comparisons between multiple outputs~\cite{direct_preference2023}. Once trained, this reward model serves as a proxy for human judgment, allowing the system to automatically score new generations and provide a reward signal for downstream RL optimization. This mechanism helps align the language model with human values and task-specific objectives, without~requiring constant human~intervention.

By analogy, in~the context of formal reasoning, one could imagine training a model to predict the success or quality of a proof attempt not by verifying it directly with a proof assistant, but~by estimating attributes such as likelihood of success, structural coherence, or~even elegance. Such a model could serve as a cheap and differentiable substitute for a proof assistant during learning. Although~it would not be logically authoritative, it could still guide exploration, enable reward shaping, and~improve sampling efficiency in RL-style~training.

In the context of mathematical reasoning, many different rewarding models have been proposed, 
both based on output evaluation and preferences~\cite{helpsteer2,AceMath}
or on step-by-step evaluations of model responses, 
as a mechanism to help discerning the optimal solution paths for multi-step tasks~\cite{Step-wise-rewards2025}.

\subsection{Reinforcement~Learning}

LLMs are frequently trained using RL techniques, usually guided by human preferences or preference models~\cite{RLHF25}. In~classical RL, an~agent interacts with an environment; at each timestep $t$, it selects an action based on the
current state $s_t$, transitions to a new state $s_{t+1}$, and~receives a reward $r_t$. The~agent's behavior is represented by a stochastic policy $\pi(a \mid s)$, and~the objective is to
learn a policy that maximizes the expected cumulative (possibly discounted) future~reward.

In the context of LLMs; however, the~notion of “environment” differs markedly from
traditional RL settings. There is no external world evolving independently in response to
actions.~Instead,
\begin{itemize}
\item The state \textls[-15]{corresponds to the current prompt together with all previously
generated tokens;}
\item The action is the next token selected and appended to the sequence.
\end{itemize}

Unlike in classical RL settings, there is no external environment that evolves
independently in response to actions. The~only evolution of the ``state'' occurs
through the autoregressive accumulation of tokens. As~each token is generated, it
extends the prompt, creating a new textual context for the next~step.

In this framing, the~model's behavior can be interpreted as learning a policy over
token sequences, deciding, at~each step, how best to continue the current trajectory
based on the accumulated history. This view underpins reinforcement learning
approaches applied to LLMs, where the quality of the generated text is evaluated
globally (for instance, by~a reward model trained from human feedback, or~by an
external verifier), and~learning adjusts the model's generation policy accordingly
~\cite{ouyang2022training}. In~practice, policy-gradient methods such as
PPO or related algorithms are commonly used to slightly shift the pretrained
distribution towards higher-reward~continuations.

Rewards are typically sparse, since they are often provided only at the end of the
episode. This sparsity poses challenges for long-horizon tasks such as coding or mathematical
reasoning, where many intermediate steps contribute to success but do not receive explicit feedback. Some examples of reward signals used for LLM training across different domains
are given in Table~\ref{tab:rewards}. Recent work has also investigated shaping these sparse
signals through structured intermediate evaluations or verifier-assisted feedback
~\cite{lyu2023faithful,Xu2024faithful}, which is particularly relevant for tasks involving multi-step~reasoning.

\begin{table}[H]
\small
\caption{Typical %MDPI: Please note that changes to the position/size of figures or tables may occur during the production stage.
 reward signals used in RL-based techniques across different LLM~scenarios.}
%	\centering
\begin{tabularx}{\textwidth}{LLL}
\toprule
\textbf{Scenario} & \textbf{Environment} & \textbf{Reward Signal} \\
\midrule
RLHF (Human Feedback) &
Human / reward model &
Which output is preferred? \\
Code/math correctness &
Program evaluator / test suite &
Did the program pass tests? \\
Proof generation &
Proof assistant &
Did the proof succeed? Was the tactic valid? \\
Format adherence &
External validator &
Did the output respect a given format? \\
\bottomrule
\end{tabularx}
\label{tab:rewards}

\end{table}
\unskip

\subsection{Inference-Time~Scaling}
Inference-time scaling refers to a broad class of techniques that improve the performance of a language model at test time, without~changing the parameters of the model or retraining it. Instead of scaling the model itself (e.g., with~more parameters or training data), inference-time scaling increases performance by investing more computation, time, or~structured reasoning during~inference.

This concept has gained importance as language models demonstrate improved capabilities when allowed to ``think longer'' or ``try harder,'' even with fixed~weights.

At its core, the~approach transforms a model's potential into actual performance by trying more options, smart filtering, iterating with feedback or delegating task to auxiliary~tools. 

Basic techniques~comprise the following:
\begin{enumerate}
\item Massive Sampling and Reranking~\cite{data_selection,code_reranker,re_ranking_chatgpt}.
Sample dozens, hundreds, or~thousands of completions per prompt (e.g., pass@k), and~then select the best using external metrics (e.g., test-case success, proof checkers), model-internal confidence or voting.
\item Iterative or Multi-Turn Inference: Refine outputs step-by-step (e.g., self-correction~\cite{self-correction}, critique-and-revise~\cite{critic2024gou}). \textls[-15]{This enables retrying or backtracking through reasoning space, incorporate feedback from prior failures (e.g., from~a proof assistant or a compiler}).
\item Tool Use and External Calls: 
Leverage calculators, theorem provers, or~web search~\cite{active_retrieval} to verify or solve subproblems, detect errors, and~possibly correct them.
This converts static model reasoning into interactive, mixed-system reasoning.
\item Latency and Budget Trade-Offs. 
The model may produce better outputs if allowed to run longer,
try more paths or search deeper. These are deliberate trade-offs between inference speed and solution quality.
\end{enumerate}

\subsection{Wait, Let us Think This~Through}
LLMs like GPT, Claude, and~others are trained on vast amounts of web text, conversations, forums, tutorials, etc., where phrases~like:
\begin{itemize}
\item “Wait, wait, wait$\ldots$”;
\item “Hold on a second$\ldots$”;
\item “Let us think carefully here$\ldots$”;
\item “Before we answer$\ldots$”.
\end{itemize}
frequently precede rethinking, correction, or~a more thoughtful explanation. So, during~pretraining, the~model picks up on these discourse~patterns.

As a result, even though models are not explicitly taught what to do when they see `wait, wait, wait' they associate it with reflection, reconsideration, or~step-by-step thinking, and~often respond~accordingly. 

These are all variants of meta-cognitive prompting, which have become a mainstream prompting strategy, especially in Chain-of-Thought~\cite{CoT,ZeroShotCoT,ToT2023} reasoning (e.g., math and logic), Coding and Proof~generation. 

Modern high-performance setups (like DeepSeek-Prover, Minerva, AlphaCode 2, or~GPT-4 Turbo in tool-augmented settings) 
typically do the “wait, wait, wait” reasoning on their own, often guided by internal reward signals, feedback loops, or~structured inference-time logic. This behavior is~as follows:
\begin{itemize}
\item Hard-coded in the inference strategy (e.g., rerun on failure);
\item Prompt-internalized (CoT triggered from the instruction);
\item Latent in system design (e.g., RAG, tool agents).
\end{itemize}

So, the~“wait, wait, wait” prompt is no longer necessary in well-designed inference-time pipelines, but~remains a useful option when prompting LLMs directly, especially in less structured or open-ended use~cases.

\section{Datasets and~Benchmarks}
\label{sec:datasets}
Many datasets exist addressing mathematical reasoning. In~this section, we discuss some of the most recent and challenging ones, frequently adopted for benchmarking models: AIME 2024, PGPS9K, miniF2F, and~FrontierMath. The~examination of these datasets and benchmarks will help us motivate our later comparison of systems and clarify why the models we focus on are particularly~relevant.

In addition to these core datasets, a~few others deserve to be mentioned.
\mbox{MATH-500} (\url{https://huggingface.co/datasets/HuggingFaceH4/MATH-500}, accessed on December 23 2025)  is a curated subset of 500 problems introduced in~\cite{MATH-500}, derived from the original MATH dataset of Hendrycks~et~al.~\cite{MATHdataset}. However, recent models such as OpenAI-o1 or DeepSeek-R1 achieve around 97\% accuracy on this benchmark, so it is now essentially saturated and unlikely to serve as a meaningful discriminator for future~systems.

Despite its comprehensive design and substantial size, the~LeanDojo dataset~\cite{LeanDojo23} has not yet achieved widespread adoption as a standard benchmark in the theorem-proving~community.

Among other interesting datasets, we also recall LILA~\cite{LILA}, expressing tasks and solutions in the form of Python
 programs, and~NumGLUE~\cite{NumGLUE}, a~multi-task benchmark combining several types of arithmetic and numeracy~challenges.

Before turning to individual benchmarks, it is important to stress that commonly reported evaluation metrics such as accuracy, pass@k, or~proof success rates are not directly comparable across different mathematical reasoning tasks. These metrics capture distinct notions of performance depending on the interaction paradigm (single-shot prediction, sampling-based generation, or~verifier-guided search), the~availability of intermediate feedback, and~the structure of the benchmark itself. Consequently, improvements within a given benchmark may reflect better adaptation to task-specific regularities or evaluation protocols rather than a general increase in mathematical reasoning ability. Throughout the remainder of this section, reported scores should therefore be interpreted as context-dependent indicators rather than absolute measures of reasoning~competence.

%Finally, a note about the figures included in this section: all performance diagrams were originally taken from the “papers with code” platform. Since that service was discontinued by Meta, the diagrams are no longer publicly accessible online; we nonetheless retain them here because they convey useful comparative information that would otherwise be difficult to reconstruct.

\subsection{AIME~2024}
The AIME 2024 dataset (\url{https://huggingface.co/datasets/Maxwell-Jia/AIME_2024}, accessed on December 23 2025) is a curated collection of 30 problems from the 2024 American Invitational Mathematics Examination (AIME), a~highly selective high school mathematics competition administered by the Mathematical Association of America (MAA) (\url{https://maa.org/}, accessed on December 23 2025). These dataset includes problems from both AIME I and AIME II, along with their official answers and detailed~solutions.

The dataset contains short, self-contained problems that require a numerical
answer between 0 and~999.  
They cover a broad range of topics comprising algebra, combinatorics, number theory, geometry, and~probability. Problems typically require multi-step reasoning or creative problem decomposition. Each entry includes an identifier, the~problem statement in natural language (with \LaTeX{}
 where appropriate),
a step-by-step solution, and~the final integer~answer.

From the perspective of mathematical reasoning, AIME problems are valuable
because they stress symbolic manipulation, quantitative reasoning, and~multi-hop inference while remaining compact enough to discourage superficial
pattern-matching. They form an intermediate difficulty tier between textbook
exercises and olympiad-level mathematics and are widely used to evaluate the
robustness of chain-of-thought~reasoning.

Current models display strong but still limited performance.  
DeepSeek-R1~\cite{DeepSeek-R1} reports an accuracy of 79.8\% on this benchmark, slightly above
general-purpose systems such as OpenAI-o1.  
The dataset is therefore not yet saturated and remains a useful testbed for
assessing improvements in systematic mathematical~reasoning.

\subsection{PGPS9K}
GPS9K (Plane Geometry Problem Solving 9K) \cite{PSPGNet} is a large-scale benchmark of the Chinese Academy of Sciences specifically designed to evaluate multimodal mathematical reasoning in the context of Euclidean~geometry. 

It comprises 9022
 geometry problems, each paired with a diagram and a set of structured annotations. The~dataset was developed to test the ability 
of the models to integrate visual understanding with symbolic reasoning, making it substantially different from purely textual mathematical~benchmarks.

The dataset contains problems spanning thirty categories of plane geometry, including angle relations, triangle congruence, circle theorems, parallel lines, similar figures, and~coordinate geometry. Each problem includes (i) a natural language description, \mbox{(ii) a diagram,} and~(iii) a collection of structural and semantic clauses describing geometric relations extracted from the diagram (for example, incidence, parallelism, angle equalities, and~metric constraints). A~formal solution program is also provided, expressed as a sequence of operations drawn from a fixed vocabulary of geometric~primitives.

PGPS9K is relevant for mathematical reasoning because geometry introduces a perceptual component absent from algebraic or number-theoretic benchmarks---solving a problem requires integrating textual statements with spatial configurations. This makes the dataset an important test of whether LLM-based systems can perform genuinely multimodal deductive reasoning, rather than relying on linguistic pattern recognition alone. The~presence of structured clause annotations also makes it possible to study how models use (or fail to use) intermediate geometric~information.

Current benchmarks indicate that geometry-specific systems significantly outperform general-purpose models. GOLD~\cite{GOLD} currently achieves the best performance, with~an accuracy of 65.8\%, followed by PGPSNet~\cite{PSPGNet}. These systems rely on a hybrid architecture 
that combines computer vision modules for diagram interpretation with LLMs for symbolic reasoning. General LLMs without specialized diagram--processing components perform considerably worse, underscoring the difficulty of integrating visual and symbolic sources of information in a coherent reasoning~pipeline.

Interest in visual and multimodal systems is growing rapidly, making this one of the most stimulating emerging research areas in mathematical reasoning~\cite{MathCoder-VL,CodePlot,MathCanvas25}.

%\begin{figure}[H]
%\label{fig:PGPS9K}
%    \includegraphics[width=\textwidth]{PGPS9K.png}
%    \caption{Diagram from \href{https://paperswithcode.com/sota/mathematical-%reasoning-on-pgps9k}{paper with code}.}
%\end{figure}

\subsection{miniF2F}
The miniF2F benchmark~\cite{miniF2F} is designed to evaluate the ability of machine learning models to solve mathematical problems expressed in natural language and formalized in the Lean theorem prover~\cite{LEAN}. Specifically,
it focuses on problems that combine traditional mathematical phrasing with rigorously specified formal statements, thereby serving as a bridge between the two~domains.

The dataset consists of problems originating from diverse sources, including mathematical olympiads (e.g., AMC and AIME), proof-based undergraduate textbooks, and~community-curated Lean theorems. For~each problem, miniF2F provides a natural language version alongside its corresponding Lean formalization; in some cases, an~initial tactic state or a reference formal proof is also included. Note that several users and developers have pointed out that
some formalizations were incorrect or ambiguous, due to misinterpretation of the original natural language problems. Even when syntactically valid, semantic mismatches sometimes existed (e.g., subtleties in quantifiers, constraints, or~domains). As~a result, different research groups often curated their own corrected subsets of miniF2F for evaluation, leading to difficulties in benchmarking and comparisons.

%MiniF2F is particularly relevant because it exposes two fundamental difficulties for LLM-based reasoning:
MiniF2F remains however particularly relevant because it exposes two fundamental difficulties for LLM-based reasoning: (i) the translation gap between human-readable mathematics and fully formalized statements, and~(ii) the need for multi-step symbolic reasoning within the syntactic constraints of an interactive theorem prover. The~benchmark tests algebra, number theory, geometry, combinatorics, inequalities, real analysis, and~elementary \mbox{calculus---domains} where informal reasoning habits often fail to align with formal \mbox{proof~requirements.}

%The chart in Figure~\ref{fig:minif2f}, borrowed form the 
%\href{https://paperswithcode.com/sota/automated-theorem-proving-on-minif2f-test}{paper with code} site, shows the evolution of the performance in recent years.
%\begin{figure}[H]
%\label{fig:minif2f}
%    \includegraphics[width=\textwidth]{chart_minif2f.png}
%\caption{Diagram from \href{https://paperswithcode.com/sota/automated-theorem-proving-on-minif2f-test}{paper with code}.}
%\end{figure}

Over the past few years, miniF2F has played a central role in tracking progress in neural theorem proving. Recent systems have achieved substantial improvements, culminating in Kimina-Prover~\cite{kiminaprover}, which currently holds the state-of-the-art with a pass@8192 score above 80\%.
We recall that pass@k is a metric used to evaluate code generation models: it measures the probability that at least one out of $k$ generated outputs solves the task correctly. This~metric is commonly used when models sample multiple candidate solutions per problem, and~a solution is considered ``passed" if it meets correctness criteria such as passing a test suite.

Other notable approaches include ProofAug~\cite{ProofAug}, DeepSeek-Prover~\cite{DeepSeek-Prover-V1.5}, Lyra~\cite{Lyra2024}  and earlier systems such as Evariste~\cite{Evariste22}, LegoProver~\cite{LEGO-Prover2024}, and~LeanCopilot~\cite{LeanCopilot24}. Despite this progress, consistent performance on more complex or less templated problems remains elusive, making miniF2F a useful stress test for formal mathematical reasoning with~LLMs.

\subsection{FrontierMath}
FrontierMath~\cite{FrontierMath} is \textls[-15]{a very recent benchmark developed by Epoch AI
 (\url{https://epoch.ai/} accessed on December 23 2025) }in collaboration with over 70 mathematicians from leading institutions. It was intentionally designed to evaluate advanced mathematical reasoning on problems far beyond standard competition or textbook level. Unlike most public datasets, FrontierMath consists of entirely novel and unpublished problems, created under strict data-protection protocols to prevent contamination. Its primary goal is to assess whether modern LLMs can handle genuinely new mathematics rather than relying on memorization or pattern~matching.

The dataset contains problems spanning a wide spectrum of high-level mathematics, including algebraic geometry, number theory, real analysis, category theory, and~set theory. These problems were written by more than seventy mathematicians invited by Epoch AI, and~organized into multiple tiers of increasing difficulty, with~the highest tier targeting challenges approachable only by experts in the respective subfields. All problems are designed to require deep multi-step reasoning and cannot be solved by brute force or heuristic~shortcuts.

FrontierMath is relevant for evaluating LLMs because it deliberately targets the limitations exposed by earlier benchmarks, many of which conflate general reasoning ability with proficiency in exploiting benchmark-specific heuristics or recurring templates. Its problems require original insight rather than template recognition, and~they provide a clean test of reasoning capabilities in settings where data leakage is nearly impossible.
The benchmark has therefore been positioned as an upper-bound stress test for the current generation of large~models.

Only a few systems have been evaluated so far. General-purpose models such as GPT-4 or Gemini achieve very low accuracy (below 2\%), while OpenAI's o3 reportedly achieves a substantially higher score, around 25\%. These numbers remain small in absolute terms, but~the large gap between o3 and previous models highlights the potential impact of improved reasoning pipelines and inference-time strategies. 
Note that OpenAI o3 result is in fact somewhat controversial, since it came to light that OpenAI had financed FrontierMath and held access to nearly the full set of problems and solutions, while Epoch AI only retained a 50-problem holdout for independent evaluation. Critics argued that contributors were unaware of OpenAI's involvement, and~questioned whether the results were free from “data contamination” due to insider access. Voices on platforms like LessWrong, Reddit, and~TechCrunch highlighted concerns over transparency, fairness, and~the integrity of benchmarks. In~response, Epoch AI and OpenAI issued public clarifications, stressing that a verbal agreement prohibited OpenAI from using the problems for training, and~emphasized that the benchmark is based on a withheld subset meant for unbiased testing.

%The conception and organization of FrontierMath has recently been presented in an \href{https://lemmata.substack.com/p/interview-with-elliot-glazer-lead}{interview}\footnote{\url{https://lemmata.substack.com/p/interview-with-elliot-glazer-lead}} with E.Glazer, the head mathematician of the project. All problems are organized into progressively more difficult tiers, to systematically evaluate AI capabilities:
%\begin{itemize}
%\item Tiers 1--3 consist of (hard) unpublished mathematics problems spanning from undergraduate to graduate/research-level difficulty.
%\item As AI performance improved, Epoch AI introduced a Tier 4, designed to challenge even experienced academic mathematicians. These problems are crafted to be truly original and difficult, often requiring novel reasoning that goes beyond standard textbook material.
%\end{itemize}
%To assemble Tier 4, Epoch AI hosted a closed symposium where mathematicians worked under non-disclosure and strict protocols (e.g., using Signal to communicate to avoid leaks) to design problems that are hard, computationally verifiable, but resistant to brute-force or guesswork.

%\begin{figure}[H]
%\label{fig:FrontierMath}
%    \includegraphics[width=\textwidth]{chart_FrontierMath.png}
%    \caption{Diagram from \href{https://paperswithcode.com/sota/mathematical-reasoning-on-frontiermath}{paper with code}.}
%\end{figure}
%Remarkably, these are general purpose systems, not specifically 
%tailored to solve mathematical problems. Given the complexity
%of FrontierMath, the fact that they can partially solve them is an 
%astonishing result.

\section{Comparative Model~Discussion}
\label{sec:comparative_discussion}
This section examines how current LLM systems behave across three environments---traditional mathematics, code synthesis, and~formal proof---in order to prepare the discussion in Section~\ref{sec:major_questions} on the three overarching~questions.

Before diving into the technical comparison of models for traditional and formalized mathematical reasoning, it is important to highlight a fundamental difference in how LLMs interact with the setting of traditional and formalized~mathematics.

In traditional mathematics, the~model is typically asked to produce a full argument in natural language, closely resembling the structure and style of human-written proofs. This is often performed in a single pass, though~additional refinement steps may occur in multi-turn or self-corrective settings. The~output is expected to follow a coherent chain of reasoning, with~intermediate steps articulated explicitly, something current state-of-the-art models do not always achieve~\cite{proof_or_bluff}.
%\footnote{\label{proofflaw}In~\cite{proof_or_bluff}, it is argued that although current models achieve performance comparable to top human competitors in giving the right numerical answer, the arguments that models produce in order to support their answer do not meet the standards of rigor that we expect from a mathematical proof. In particular, the authors of this paper asked human evaluators ``having substantial mathematical problem-solving experience as former national IMO team members or having participated in final-stage team selection processes for their countries'' \cite[p.~2]{proof_or_bluff}, to evaluate the arguments produced by LLMs to justify the numerical results they generated. According to these experts, the arguments produced by LLMs contain four main kinds of flaws: (i) errors due to logical fallacies or unjustified reasoning steps; (ii) errors coming from the introduction of unproven or incorrect assumptions; (iii) errors resulting from fundamentally incorrect solution strategies; (iv) errors concerning algebraic or arithmetic miscalculations.}  
If the task demands a final numerical answer, as~in many competition-style problems, the~model is also expected to isolate and present the result clearly at the end of the~derivation.

This style aligns naturally with the capabilities of LLMs. It mirrors the material encountered during pretraining---textbooks, lecture notes, and~mathematical discussions---making the generation task relatively familiar. Chain-of-thought prompting further improves performance on complex reasoning tasks by encouraging a step-by-step derivation. Many evaluation protocols for traditional benchmarks reinforce this behavior by assessing not only correctness but also clarity and validity of the reasoning~process.

In contrast, formal mathematics adopts a very different interaction paradigm. Rather than generating full proofs, the~model typically produces one tactic or inference step at a time. Each tactic is passed to a proof assistant, who checks its syntactic and logical correctness and updates the internal proof state accordingly. This updated state is then fed back into the model as context for the next prediction. The~resulting process is a fine-grained, iterative loop in which the proof is constructed incrementally and under the supervision of a formal~system.

% These contrasting interaction patterns---narrative generation versus step-by-step tactical inference---highlight a key divergence in how LLMs engage with traditional and formalized mathematics.

A summary of these differences is provided in Table~\ref{tab:informal_vs_formal_math}.

% is worth noting that while the step-by-step interaction fits naturally with the architecture of interactive theorem provers, it should not be mistaken for a principled methodological choice. Rather, it reflects the current state-of-the-art and the practical constraints imposed by existing tools. There is no inherent reason why formal proofs could not be approached at a coarser level of granularity, especially if adopting a declarative proof style instead of a procedural one. We shall come back on this point in Section~\ref{sec:LLMs_formal_vs_informal}.

\subsection{Traditional Mathematical~Reasoning}
\label{sec:traditional_models}
General-purpose LLMs such as OpenAI's o3 and o4 models or Claude~3.5 already
exhibit strong performance on many traditional mathematical benchmarks. This is not surprising, the task resembles the material encountered during pretraining---textbook derivations, lecture notes, forum discussions---so generating a full chain of reasoning in natural language largely amounts to reproducing familiar discourse patterns. Math-specific fine-tuning further reinforces this behavior by providing explicit demonstrations of structured reasoning, coherent step-by-step argumentation, and~standard \mbox{mathematical~formatting.}

\begin{table}[t] %This table belongs to the previous section! Do not use [H]
%\centering
\caption{Comparison 
 of traditional vs. formalized mathematical reasoning with~LLMs.}
\begin{tabularx}{\textwidth}{lLL}
\toprule
\textbf{Aspect} & \textbf{Traditional Mathematics} & \textbf{Formalized Mathematics} \\
\midrule
Language 
& Natural language with embedded math notation & Rigid symbolic language enforced by proof assistant \\
\midrule
Output Granularity & Full argument or solution in one pass & Single tactic or step at a time \\
\midrule
Feedback Style & Sparse; often based on final answer correctness or \linebreak  human judgment & Frequent; each step checked by proof assistant for validity \\
\midrule
Proof State & Implicit; carried in the text or inferred from structure & Explicit; updated after every tactic via proof assistant \\
\midrule
Goal Type & Plausibility, clarity, coherence & Formal logical correctness \\
\midrule
Model Role & Autonomous generator of human-style solutions & Step-wise assistant guided by tool feedback \\
\midrule
Training Supervision & Coarse-grained; from textbooks, forums, solutions & Fine-grained; tactic-state pairs, often RL-based \\
\midrule
Tool Integration & Optional or post hoc checking & Tight integration with ITP systems like Lean, Coq \\
\bottomrule
\end{tabularx}
\label{tab:informal_vs_formal_math}
\end{table}

Among recent systems designed for mathematical problem-solving in natural language,
two models stand out as paradigmatic examples of contrasting design philosophies: Minerva (Google 2022) \cite{Minerva22, Minerva23} and DeepSeek-R1 (DeepSeek-AI, 2024) \cite{DeepSeek-R1}. Although~both
achieve strong performance on competition-style problems, they represent almost opposite
approaches to inducing mathematical reasoning in LLMs. The~differences between the two systems are summarized in Table~\ref{tab:minerva_vs_deepseek}, and~discussed in more detail~below.

Minerva is the canonical instance of a supervised imitation paradigm. The~model
is pretrained at a very large scale (PaLM~540B) and then shaped through extensive
supervised fine-tuning on high-quality chain-of-thought solutions drawn from ArXiv, textbooks, and~math-heavy sources. Reasoning is therefore learned by example: the model internalizes the structure of human-written derivations and relies heavily on sampling methods such as self-consistency (pass@k) to select coherent solutions. This approach leverages scale and high-quality supervision, producing strong results, but~depends critically on the availability and correctness of detailed solution~traces.

DeepSeek-R1 follows a markedly different philosophy. Instead of supervised reasoning
trajectories, the~model starts from a math-oriented base model
(DeepSeekMath-Base~\cite{DeepSeekMath}) and is trained entirely through RL
guided by a reward model~\cite{DeepSeek-Prover-V1.5}. Chain-of-thought labels are not used;
the only learning signal comes from outcome-based rewards. This cold-start strategy ---i.e., a strategy based on a few thousand of carefully curated problems with structured formats, but without step-by-step reasoning labels, serving as initial scaffolding for reward shaping---
is meant to encourage the model to develop its own reasoning heuristics rather
than imitate human-derived patterns. Although~this approach reduces reliance on labeled data and can yield more exploratory reasoning behavior, it also faces the well-known instabilities of RL, particularly in sparse-reward~settings.

These two models illuminate complementary strengths and limitations. Minerva shows what can be achieved through scale and high-quality supervision; DeepSeek-R1, through structured exploration and reward-driven reasoning. Both highlight that strong performance in mathematical problem-solving depends not only on model size and training data, but~also on how reasoning is induced, stabilized, and~postprocessed.

\begin{table}[H]
\small
    \caption{Comparison between Minerva and DeepSeek-R1. Minerva adopts a supervised imitation paradigm, while DeepSeek-R1 is entirely trained though Reinforcement~Learning.}
%    \centering
    \begin{tabularx}{\textwidth}{lLL}
    \toprule
         \textbf{Aspect} & \textbf{Minerva (Google, 2022)} & \textbf{DeepSeek-R1 \linebreak  (DeepSeek-AI, 2024)}\\\midrule
         Base Model & PaLM 540B & DeepSeekMath-Base \linebreak  (custom pretraining) \\\midrule
         Training Data & ArXiv papers, textbook problems, math-heavy Wikipedia & DeepSeekMath corpus + curated cold-start problems \\\midrule
         Supervised Fine-Tuning & Yes---CoT-style solutions & No---RL only \\\midrule
         RL/Reward Model & None & PPO-style RL guided by learned reward model \\\midrule
         Cold-Start Strategy & Not used & Yes---no reasoning labels \\\midrule
         Reasoning Supervision & CoT with full solutions & Only reward signals \\\midrule
         Representation Format & Natural language + LaTeX & Natural language with internal reasoning steps \\\midrule
         Postprocessing & Self-consistency (pass@k) & Sampling + reward-model reranking \\\midrule
         Design Strength & Scale + high-quality supervision & Autonomous reasoning development through RL \\\midrule
         Limitation & Depends on curated \linebreak  reasoning traces & RL instability; sparse reward \\
         \bottomrule
    \end{tabularx}
    \label{tab:minerva_vs_deepseek}
\end{table}

\subsection{Formalized Mathematical~Reasoning}
\label{subsec:formal_math}

In formal mathematics, the~interaction paradigm is entirely different from the 
traditional setting. Instead of producing a full proof in one pass, the~model 
generates a single tactic or inference step at a time. This step is passed 
to a proof assistant, which checks its correctness and updates the proof state. 
The updated state is then fed back to the model as part of the next prompt. 
Proof construction proceeds as a fine-grained loop of prediction, verification, 
and state~update.

This structure reflects the architecture of interactive theorem provers rather 
than a deliberate methodological choice. There is no intrinsic reason why formal 
proof generation must proceed via such atomic steps; in principle, one could 
instead target larger proof fragments or declarative chunks. However, current 
proof assistants expose their APIs at the tactic level, and~contemporary systems 
have been designed accordingly. The~consequence is that the model never learns a 
persistent notion of proof state internally; instead, the~state remains external 
and is reintroduced at every~iteration.

\begin{figure}[H]
%    \centering
    \includegraphics[width=0.6\linewidth]{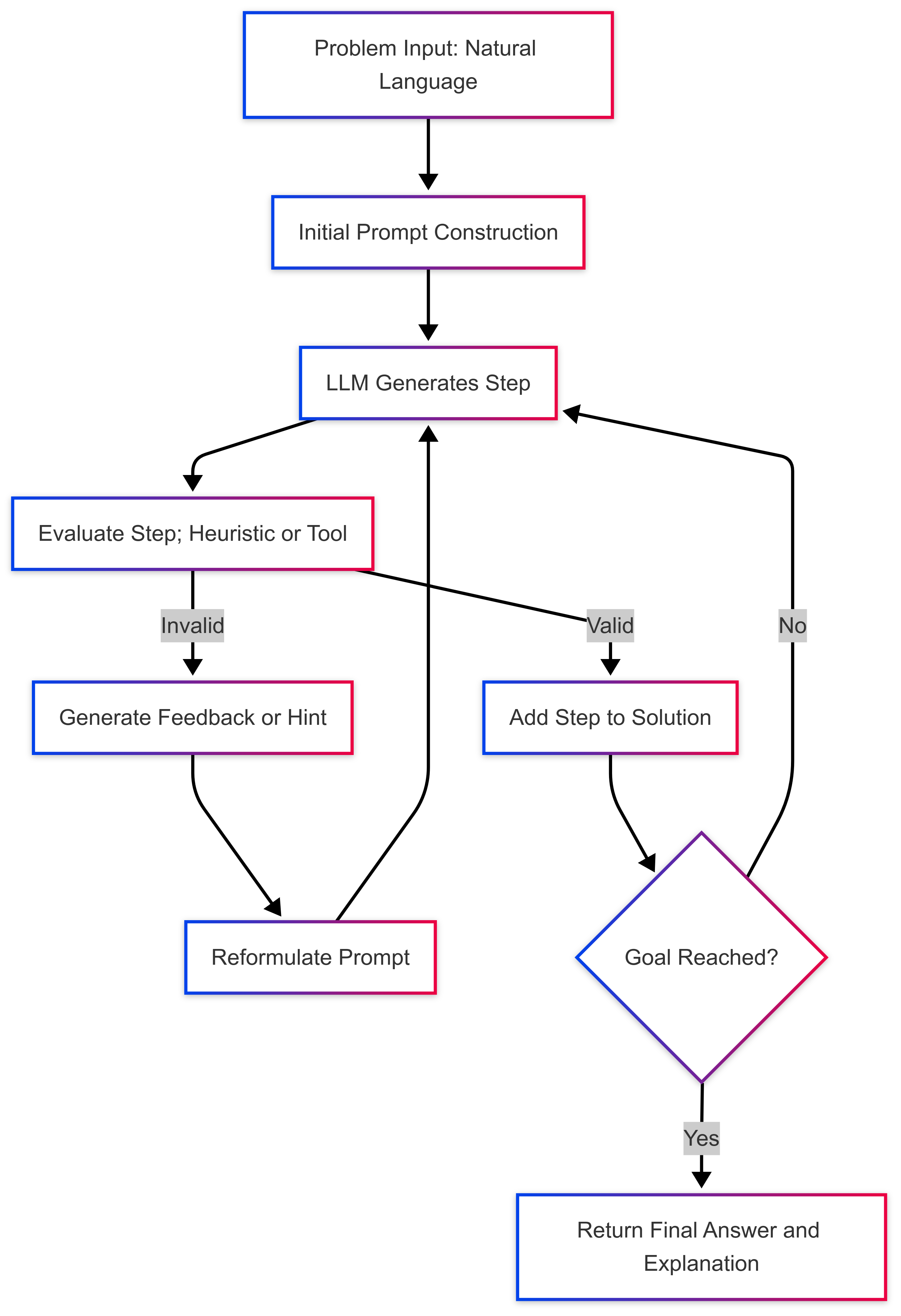}
    \caption{Typical 
 interaction loop between a formal prover and a Large Language Model is the case of formal mathematics. The~LLM proposes a proof step (e.g., a~tactic), which is checked by an external proof assistant that explicitly maintains and updates the proof state. Unlike standard reinforcement learning loops with an evolving external environment, state transitions here are symbolic and tool-mediated, and~the model does not internally persist the deductive state but receives it anew at each iteration. Feedback or hints are generated when necessary, and~the loop continues until a complete proof is~assembled.}
    \label{fig:revision_loop}
\end{figure}

Figure~\ref{fig:revision_loop} summarizes this interaction pattern in the 
context of recent systems. The~model proposes a step, a~verifier or heuristic 
evaluates it, and~the environment provides structured feedback that shapes the 
next prediction. This is representative of systems such as Kimina-Prover, Lyra, 
and DeepSeek-Prover, although~each implements its own variants of state 
representation, error handling, and~prompt~construction.

Although effective, this mode of interaction also exposes the limitations of current 
LLMs with respect to mathematical reasoning. Maintaining global proof structure, 
tracking dependencies, or~pursuing long-term strategies is offloaded to the proof 
assistant, interrupting the model's ability to develop internal mechanisms for 
state tracking. As~we shall see in the next subsections, different models 
respond to these constraints differently---some rely on extensive sampling and 
reranking, others use reward-based refinement, and~some attempt to compress the 
proof state into the prompt. These differences highlight the gap between 
performing mathematics externally through tools and learning to reason 
internally in a manner closer to mathematical~practice.

In the next sections, we will analyze the following more~detail:
\begin{itemize}
\item How the context is defined, represented, and~retrieved.
\item How the proof trajectory supports reinforcement learning, including the role and limitations of the feedback signal.
\end{itemize}

We shall exemplify the discussion on three paradigmatic systems:
Kimina-Prover~\cite{kiminaprover}, a~recent RL-based prover built on 
Qwen2.5-72B, designed around structured proof generation in Lean~4;  
Lyra \cite{Lyra2024} that adopts a complementary philosophy, 
focusing on 
post hoc correction of model errors via tool- and conjecture-level repair 
mechanisms rather than exploration; 
DeepSeek-Prover~\cite{DeepSeek-Prover-V1.5}, combining supervised fine-tuning with 
reinforcement learning from proof-assistant feedback, together with a 
tree-search component (RMaxTS) that guides proof~exploration.  

\subsubsection{Library Retrieval and Context~Representation}
A central challenge in formalized mathematics is that proofs rely heavily on existing lemmas and definitions, making the model's ability to retrieve and represent relevant library content crucial for any non-trivial~task.

The formal prover must have access to the formal library
of the proof assistant. It is difficult to fine tune the model
over this bulk of knowledge, so provers typically rely on 
a retrieval phase. This retrieval process is purely textual: the model does not internalize the proof assistant's library as a structured, searchable object but merely consumes selected fragments as additional prompt tokens. Thus, the “context” of the model remains a linear string rather than a semantic state, and~the burden of selecting and organizing relevant information lies entirely in the retrieval pipeline. 
Since we need to retrieve theorems and
not documents, we cannot rely on available retrievers, 
but we use a system-ad hoc retriever. Due to prompt length limits, retrieval is typically
confined to top-k premises (Kimina, DeepSeek), possibly using preselection, filtering, or~dynamic selection (DeepSeek).
Both Local hypotheses and global statements are explicitly encoded in the prompt. Table~\ref{tab:retrieval}
provides a compact comparison of premise retrieval and context-injection strategies across Kimina-Prover, Lyra, and~DeepSeek-Prover, which we analyze in the following~paragraphs.

\begin{table}[H]
\small
    \centering
    \caption{Comparison
 of the library retrieval strategy between Kimina, Lyra, and~DeepSeek-Prover.}
    \begin{tabularx}{\textwidth}{cCCCC}
    \toprule
     \textbf{Model} & \textbf{Goal Representation} & \textbf{Premise Retrieval Strategy}          & \textbf{Injected into Model}  & \textbf{Notes}  \\
     \midrule              
 Kimina & Lean proof state + hypotheses & Top-k retrieved theorems via retriever & Prepended in prompt (as preamble to proof) & Retrieval is static per problem (no search-time updates) \\\midrule
  Lyra  & Lean proof state             & No dynamic retrieval of theorems       & Context = local goal state + tactic library    & Emphasis is on tool use correction, not premise lookup \\\midrule
 DeepSeek & Lean proof state + hypotheses   & Dynamic premise selection + RMaxTS  &  Retrieved theorems passed in as part of prompt & Premise selection may evolve during tree search \\\bottomrule
    \end{tabularx}
    \label{tab:retrieval}
\end{table}

In Kimina-Prover, retrieval is used to collect a fixed number of relevant theorems before proof generation. These are added to the prompt so that the model can condition on them when proposing~tactics.

Lyra does not perform premise retrieval. The~model focuses on applying correct tactics to solve the current goal. If~a tactic is invalid, correction is applied. The~relevant context is the local goal plus a known (implicit) set of tools, not imported~theorems.

DeepSeek-Prover uses premise selection (retrieval) as part of a broader search procedure (RMaxTS). The~premises supplied to the model may vary as proof branches expand, especially when recursively invoking~subproofs.

\subsubsection{Feedback and~Supervision}

All systems for formalized mathematics rely on some form of feedback from the proof
assistant. This feedback plays a central role in shaping the model's behavior, but~its
granularity and function vary substantially across different approaches.
The differences in supervision signals and feedback granularity summarized in Table~\ref{tab:feedback_supervision} underpin the contrasting learning dynamics discussed~below.

Kimina-Prover employs a reinforcement-learning pipeline in which the proof assistant
acts as a binary verifier---a proof attempt is either accepted or rejected. The~resulting
reward signal is strongly aligned with the end goal, but~highly sparse, making credit
assignment difficult over long~trajectories.

Lyra takes a different approach. Instead of exploration or RL, it relies on deterministic
post-processing guided by error messages from the proof assistant. Two mechanisms are
central: Tool Correction, which repairs invalid or hallucinated tactic calls, and~Conjecture Correction, which adjusts subgoals in response to unprovable or
ill-formed statements. This yields excellent sample efficiency, but~offers limited opportunity for models to learn new proof~strategies.

\begin{table}[H]
    \small
    \caption{Comparison
 between Kimina-Prover, Lyra, and~DeepSeek-Prover with respect to supervision and~feedback.}
%    \centering
    \begin{tabularx}{\textwidth}{cCCC}
    \toprule
    \textbf{Model} & \textbf{Supervision Type} &
    \textbf{Feedback Source} & \textbf{Feedback Granularity} \\
    \midrule
    Kimina-Prover &
    Reinforcement Learning &
    Lean 4 verifier &
    Whole-proof success/failure (pass/fail) \\
    \midrule
    Lyra &
    Supervised + error-driven correction &
    Proof assistant error messages &
    Local tool-level and goal-level feedback \\
    \midrule
    DeepSeek-Prover &
    Supervised Fine-Tuning + RLPAF &
    Lean runtime (tactic acceptance) &
    Step-by-step tactic validation \\
    \bottomrule
    \end{tabularx}
    \label{tab:feedback_supervision}
\end{table}

DeepSeek-Prover integrates supervision more tightly. After~a supervised fine-tuning
stage on formal proof data, it undergoes reinforcement learning from proof assistant
feedback (RLPAF), where each tactic is executed and checked in real time. This produces a far denser reward signal than whole-proof verification and enables the model to refine
its behavior at the granularity of single inference~steps.

Overall, the~three systems illustrate a spectrum of feedback regimes---from sparse but
highly goal-aligned (Kimina) to dense and execution-level (DeepSeek) to deterministic
corrective supervision (Lyra)---each reflecting a different philosophy about how formal
reasoning should be learned and~controlled.

These observations set up the analysis in Sections~\ref{sec:autoformalization} and \ref{sec:major_questions}, where we discuss formalization explicitly and relate these empirical patterns to the three major questions introduced in the~Introduction.

\section{Autoformalization}
\label{sec:autoformalization}
The following Section connects the empirical observations from 
Section~\ref{sec:comparative_discussion} to the question of how traditional and formalized mathematics differ as learning targets. In~particular, autoformalization illuminates where current models struggle when transitioning from narrative reasoning to formal symbolic~structures.

In Section~\ref{sec:informalformal}, we sketched a distinction between traditional and
formalized mathematical reasoning, while standard mathematical practice is still largely
conducted in the traditional setting, there is a growing interest in the formalization of
mathematics. In~particular, we are witnessing the emergence of educational and research
programs that promote the idea that mathematical texts, especially proofs, should be
written in a fully formalized manner, allowing for automated verification by computers
~\cite{QEDrev07, Voevodsky14, Buzzard24, Avigad24, HLK24}.

Behind such programs lies the idea that a mathematical text can, in~principle, always be
translated into a formal one. In the domain of mathematical logic, this idea is dubbed by some authors as the ``Hilbert's Thesis'' \cite[p.~41]{Barwise77}\cite[p.~11]{Rav99}, while other authors call it the ``standard view of proofs'' \cite{AntonuttiMarfori10}.
Well-known examples corroborating this view include the formalization of the proof of the
Kepler conjecture~\cite{HalesKeplerConjecture}, the~Four-Color theorem~\cite{Gonthier07},
the Feit--Thompson theorem~\cite{Gonthier13}, and~the ongoing project of formalizing the
proof of Fermat's Last Theorem~\cite{fermat_last_theorem}. These formalizations were
performed manually and required years of work, involving the translation of extensive
fragments of mathematics (definitions, axioms, lemmas, derived constructions, and~so on).

Would it be possible to automatize such a long and sometimes tedious translation
work? This question is not new. Early, isolated attempts appeared between the late 1980s
and early 2000s (see refs.~\cite{Simon88, Simon90, Zinn04}). More systematic experiments emerged in the
last decade, motivated by advances in natural language processing and, especially,
neural-network-based translation between natural languages. This has led to the design of
autoformalization systems, i.e.,~AI systems which automatically transform a text written in
natural language mathematics into a text in formal~mathematics.

Autoformalization is particularly relevant
for AI-generated proofs: when a model outputs a derivation in natural language, an~effective autoformalization pipeline would allow one to verify whether the reasoning is
correct or~not. 

Recent work has explored this idea with increasingly ambitious methods, ranging from
sequence-to-sequence models for formal translation
~\cite{autoWu22,Lample20} to systems combining LLMs with external verification
~\cite{dont_trust24}. 

Before diving deeper into the issue of autoformalization, and~in order to better organize the discussion, it is useful to distinguish two orthogonal sources of difficulty. The~first concerns translation difficulty, namely mapping natural language mathematical expressions into the rigid syntax and vocabulary of a formal system. The~second concerns mathematical reconstruction difficulty, namely recovering implicit assumptions, missing logical steps, and~background lemmas that are routinely omitted in informal proofs. Existing systems address these challenges to varying degrees, and~their limitations often stem from different combinations of the~two.

\subsection{Early Experiments with Neural~Networks}

The first systematic attempts to train neural networks to translate traditional mathematics into formalized mathematics were made nearly a decade ago by Cezary Kaliszyk, Josef Urban, and~collaborators~\cite{KaliszykUV15, KaliszykUV17, WangKU18}. At~that time, no large corpus aligned human-written mathematical texts with their formal counterparts. The~only available resource, the~Flyspeck project, contained a few hundred paired \LaTeX{} and HOL~Light sentences---far too few for neural~training.

To overcome this limitation, the~authors generated their own synthetic data through a process they called informalization, namely the automatic transformation of formal proofs into quasi-natural forms. For~instance, in~HOL~Light, expressions like \verb+vector_add u v+ were converted into $u + v$ by removing type annotations, brackets, and~other formal syntax. The~resulting texts were more readable and ambiguous, resembling natural language while remaining far from genuine human~mathematics.

They later adopted a more linguistically expressive strategy using Mizar libraries, where theorems and proof statements already written in a semi-natural formalism were converted into ``artificial'' \LaTeX{} sentences~\cite{Bancerek06}. This yielded over one million aligned Mizar--\LaTeX{} pairs, used to train sequence-to-sequence (seq2seq) neural models based on multi-layer RNNs. Their best configuration achieved $65.7\%$ accuracy in reconstructing Mizar formulas, although~performance did not scale consistently with model~size.

While these studies were seminal in demonstrating the feasibility of autoformalization, their focus was primarily technical---the natural language side of the data were synthetically generated, limiting insights into real human--machine interaction and the linguistic variability characteristic of authentic mathematical~writing. 

These early experiments demonstrate the feasibility of autoformalization, but~they rely heavily on synthetic data and limited linguistic variability. Modern LLM-based approaches move beyond these constraints, enabling translation of real mathematical statements and definitions, which we discuss~next.

\subsection{Autoformalization of Statements and Definitions with~LLMs}

A first systematic attempt to use large language models for the autoformalization of mathematical statements is presented in~\cite{autoWu22}, exploiting the miniF2F dataset described earlier. This is a natural choice, as~miniF2F already includes competition-level math problems manually aligned with formal statements across three different proof assistants. The~models evaluated in~\cite{autoWu22} are PaLM and Codex. The~authors focus on a subset of miniF2F consisting of 140 algebra problems and 120 number theory problems, and~prompt the models to generate corresponding Isabelle formalizations. A~few-shot approach is used---10 problems are selected for prompting, while the remainder are used for evaluation. Performance is measured using BLEU~scores.

The results show that scaling up the size of models improves performance (e.g., PaLM 8B vs.\ PaLM 540B), and~that Codex outperforms PaLM overall. The~authors hypothesize that Codex's superior performance may stem from its exposure to more formal content during training. To~gain a deeper understanding of Codex's autoformalization abilities, they manually analyze its output on 150 problems from the MATH dataset. Codex produces correct formalizations for 38 of them, with~a success rate of 25.3\%. One key challenge identified is the semantic gap between natural language mathematical expressions and their formal counterparts. For~instance, when the input refers to ``the greatest possible value'', Codex often fails to map it to Isabelle's formal construct like \texttt{Greatest} or \texttt{Max}, highlighting the difficulty of aligning traditional mathematical phrasing with formal~definitions.

The problem of autoformalization of definitions has recently been addressed in~\cite{AutoFormalization}. As~stated by the authors, definitions are indeed ``a critical component of mathematical discourse'', and~although they ``serve as foundational building blocks in mathematical reasoning, yet they are often intricate, context-dependent, and~difficult to formalize'' \mbox{(\cite{AutoFormalization}, p.~2).} The~authors also take Isabelle/HOL as the target formal language, but~instead of considering miniF2F, they introduce two novel datasets: Def\_Wiki---definitions taken from Wikipedia---and Def\_ArXiv---definitions taken from research papers on arXiv. They also consider different models with respect to those used in~\cite{autoWu22}, namely GPT-4o, Llama~3, and~DeepSeek-Math.

Among the various experiments conducted, one particularly interesting direction is the investigation of the models' self-correction capabilities. The~authors explore this by feeding back into the model the formalization errors identified by the supporting proof assistant, allowing it to revise and improve its output. Another key issue they address is the challenge of formalizing definitions that require access to external formal mathematical libraries. To~mitigate this, they examine the impact of providing contextual elements---such as relevant definitions or lemmas from these libraries---as auxiliary premises to guide the model's formalization~process.

Returning to~\cite{autoWu22}, another noteworthy experiment conducted by the authors involves feeding neural theorem provers with problem statements obtained via autoformalization, and~subsequently evaluating their performance on the miniF2F benchmark. The~result is a modest but measurable improvement in the proof success rate, from~29.6\% to 32.5\%. This outcome may be interpreted as evidence that autoformalization plays a meaningful role in facilitating human--machine interaction. The~improvement suggests that the AI system benefits from translating mathematical problems authored by humans in natural language into formalized representations. In~other words, the~system performs better when operating directly on formal~expressions

A recent line of work further strengthens this perspective by introducing reflective autoformalization frameworks such as ReForm~\cite{ReForm25}, which tightly integrate semantic consistency checks into the translation loop. Rather than treating autoformalization as a direct sequence-to-sequence task, ReForm iteratively evaluates the semantic fidelity of its own formalizations and self-corrects through progressive refinement. This reflective structure parallels the broader class of iterative reasoning and self-correction techniques that we will return to in Section~\ref{sec:iterative_loops}.

\subsection{Autoformalization of~Proofs}

So far, we have considered the autoformalization of mathematical statements and definitions. Since proofs constitute the core of mathematical practice, it is natural to ask whether current systems can also transform human-written mathematical arguments into fully formal proofs. Several works explore this question, though~progress remains limited compared to autoformalizing~statements.

\subsubsection{Autoformalization of Proofs of Elementary Arithmetic and of Code Correctness in~Coq}

In~\cite{CBJ22}, the~authors investigate the autoformalization of both elementary arithmetic proofs and program--correctness proofs targeting the Coq proof assistant (nowadays known as Rocq) . Their experiments rely on two newly constructed datasets aligning informal (LaTeX-like) statements and proofs with their formal Coq~counterparts.

The first dataset consists of several families of automatically generated arithmetic statements and proofs inspired by IMO-style reasoning, covering parity, compositeness, and power relations, with a few thousand examples per family derived from reusable templates.

%The first dataset consists of three families of artificially generated arithmetic theorems and~proofs:
%\begin{itemize}
%\item even-odd: parity of algebraic expressions;
%\item \textsc{\hl{composites}}: statements asserting that a number is composite;
%\item \textsc{\hl{powers}}: statements asserting that a number is an integer power of $n$.
%\end{itemize}
%Each family contains several thousand paired statement--proof examples (\hl{5000} for \textsc{\hl{even-odd}} and \textsc{\hl{composites}}, \hl{2000} for \textsc{\hl{powers}}), produced by systematically instantiating a small set of high-level reasoning~templates.

%The second dataset, \textsc{poly}, contains \hl{5000} %MDPI: Commas are only used for numbers with five or more digits. We have removed them in four-digit numbers. Please confirm.
% examples pairing natural language descriptions of short Imp programs~\cite{Pierceetal18} with informal Hoare-style correctness proofs, each aligned to a corresponding formal Coq~script.

 The second dataset consists of 5000 examples pairing natural language descriptions of short programs written in Imp \cite{Pierceetal18}, a minimal imperative programming language commonly used in program semantics, with informal Hoare-logic correctness arguments and their corresponding formal proofs expressed in Coq.

%The model used is an encoder--decoder architecture based on the Universal Transformer~\cite{Deghani18}. Generalization is evaluated by training on expressions or programs up to certain sizes (e.g., $n \in \{2,3,5,7,9\}$) and testing on unseen sizes up to $n=12$ for arithmetic and $n=14$ for program length.

Performance is evaluated using:
(i) sequence-level accuracy, requiring an exact match with the target Coq script, and~ 
(ii) semantic accuracy, requiring only that Coq accepts the generated~proof.

The results show high semantic accuracy (over 90\%) for arithmetic proofs at moderate sizes, but~a rapid degradation for larger expressions. Program--correctness proofs are more fragile, in which accuracy reaches about 80\% for short programs but drops sharply for longer ones, failing entirely beyond length~12. A~small evaluation on 45 human-written LaTeX proof pairs yields moderate success rates, with~most errors arising from variable mismanagement or omissions caused by the rigid structure of the synthetic training~data.

Overall, these results suggest that autoformalization is feasible for simple, highly regular proof structures, but~remains brittle for more diverse or linguistically varied~arguments.

\subsubsection{Guiding Automated Theorem Provers with Informal~Proofs}

A different strategy is explored in~\cite{Minerva23}, where the goal is not to translate an informal proof directly into a full formal derivation, but~to generate a formal proof sketch that can be refined by automated theorem provers. As~emphasized in~\cite{Wiedijk03}, proof sketches preserve the high-level structure of a proof while omitting low-level details, making them suitable for refinement by formal~systems.

The target system in~\cite{Minerva23} is Isabelle (specifically, Isabelle/HOL). The LLM (Codex) generates proof sketches from informal proofs taken from miniF2F. A~few-shot prompting setup is used, in which 20 hand-crafted examples of informal proofs paired with their formal sketches serve as demonstrations. Open conjectures in the produced sketches are then discharged using Sledgehammer~\cite{Paulson10}, Isabelle tactics, or~the neural prover Thor~\cite{JiangLTCOMWJ22}.

Experiments on the 244 validation and 244 test problems of miniF2F show that this hybrid approach substantially improves proof success rates. On~the test set, Sledgehammer with tactics achieves 20.9\% success, and~Thor 29.9\%, whereas the sketch-based pipeline reaches 39.9\%. Using machine-generated informal proofs (Minerva-8B, -62B, and -540B) yields similar results: on the test set, Minerva-62B attains 37.7\% and Minerva-540B 38.9\%, close to the 39.9\% obtained from human~proofs.

This approach highlights two points. First, autoformalization need not be an all-or-nothing task---producing a structured high-level sketch can be far more tractable than producing a complete proof. Second, combining LLM-generated structure with automated theorem provers allows the system to offload low-level reasoning to symbolic tools, mitigating the length and context limitations of current~models.

\subsection{Discussion}
\label{sec:disc}

The experiments reviewed in this section highlight both the promise and the limitations of current autoformalization methods. On~the positive side, LLMs have demonstrated non-trivial abilities to translate mathematical statements, definitions, and~even fragments of proofs into formal languages. These results, obtained across different systems and proof assistants (Isabelle, Coq, Lean), indicate that the long-standing idea that traditional mathematical discourse can be systematically connected to formal representations is no longer speculative. At~least for problems of moderate linguistic and structural complexity, neural models are now able to produce formalizations accurate enough to support downstream tasks such as automated proving or proof-sketch~refinement.

At the same time, the~limitations of existing systems are equally revealing. Performance drops sharply when the linguistic form of the input deviates from patterns seen during training, when definitions depend on subtle contextual cues, or~when proofs require long chains of dependent reasoning steps. These difficulties point to two distinct but intertwined sources of complexity in autoformalization. Translating mathematical text is not merely a syntactic task. 
It requires identifying which concepts are being invoked, in~which sense they are meant, and~how they connect to the background library and logical framework of the target system. At~this translation level, failures typically manifest as syntactically valid but semantically misaligned formalizations, for~instance through inappropriate definition choices, misplaced quantifiers, or~subtle scope errors. A~second source of difficulty arises at the level of proof construction, where errors frequently stem from missing implicit steps, inappropriate intermediate lemma selection, or~breakdowns in long-range dependency management, even when the surrounding statement-level translation appears locally correct. Current models do not possess this kind of semantic situational awareness. Their success depends heavily on curated examples, explicit prompting structures, and~access to auxiliary library snippets for disambiguation~\cite{AutoFormalization,dont_trust24}.

A recurring theme across all approaches is the tension between automation and understanding. Even when an LLM-generated formalization is accepted by a proof assistant, it does not follow that the model has captured the underlying mathematical ideas. This raises the epistemological concern central to the philosophy of mathematical practice that mechanical correctness does not automatically translate into human understanding or justification. As~formalization experts have long emphasized~\cite{Wiedijk03,seventeen_provers}, the~act of formalization plays a methodological role: it forces one to analyze which assumptions are being used, how arguments decompose into elementary steps, and~whether certain commitments are necessary. Autoformalization risks bypassing this reflective process~entirely.

A further complication arises from the foundational constraints of proof assistants. Choosing the logical setting, the~library, and~the appropriate definitions is itself a delicate mathematical judgment. Current autoformalization pipelines operate within fixed frameworks and lack the capacity to reason about such choices. This is visible in the persistent failure modes of definition translation, concept selection, and~implicit assumption handling, noted both in early experiments~\cite{KaliszykUV17,WangKU18} and in recent large-scale studies~\cite{autoWu22,AutoFormalization}.

Overall, the~present state of autoformalization illustrates a mixed picture. On~the one hand, LLMs now provide practical tools for assisting formalization at scale, improving data pipelines for automated provers, and~enabling hybrid workflows combining neural and symbolic reasoning~\cite{autoWu22,Lample20}. On~the other hand, significant obstacles remain before such systems can autonomously handle the full expressive richness of mathematical text. These obstacles are not merely technical but conceptual: they stem from the gap between linguistic imitation and genuine mathematical comprehension. Addressing this gap will likely require architectures or training regimes that can internalize notions of context, state, and~higher-level structure, beyond~what current sequence models~support.

This tension between capability and limitation clarify why formal mathematics imposes qualitatively different cognitive and computational demands,
guiding the analysis \mbox{in Section~\ref{sec:major_questions}.}

\section{Major~Questions}
\label{sec:major_questions}
We now return to the three overarching questions posed in the~Introduction Section:
\begin{enumerate}
    \item To what extent are LLMs more naturally suited to traditional as opposed to \\formalized mathematics?
    \item Why is proving harder than coding?
    \item Do LLMs possess, or~approximate, a~notion of computational state relevant for coding and proving?
\end{enumerate}

Rather than addressing them in this order, we begin with the second question. Understanding why proving is harder than coding sheds light on both the differential performance of current systems and the deeper issue of whether LLMs can maintain the kinds of structured state required for mathematical~reasoning.

\subsection{Why Is Proving Harder than Coding?}
\label{sec:proving_vs_coding}
The proof assistant community is well accustomed to the Curry--Howard correspondence
which emphasizes the analogy between proving and programming (see, e.g.,~\cite{Curry_Howard_lectures}). 
Even from the perspective of large language models (LLMs), the~two domains share clear 
structural affinities: both are highly constrained, logic-driven tasks, sensitive to syntax and 
amenable to verifier-in-the-loop training. From~this viewpoint, one might expect that 
successes achieved in coding, where modern systems such as Codex or AlphaCode perform remarkably well, should naturally transfer to proving. Yet they do~not.

The standard explanation points to training data include code LLMs benefit from billions of diverse 
examples with natural language comments, test suites, and~rich redundancy, whereas formal 
proof corpora are several orders of magnitude smaller, domain-specific, and~structurally rigid. 
However, even with larger proof datasets, the~gap would remain. 
%The core difficulty lies not merely in data availability, but in how models encounter, process, and learn from errors.
This also affects how errors are surfaced during~training.

Specifically, a~useful contrast emerges when~considering the following:
\begin{itemize}
    \item What kinds of errors are exposed during generation;
    \item How those errors can be tolerated or repaired;
    \item And how models and toolchains respond to them.
\end{itemize}

\begin{figure}[H]
\centering
\includegraphics[width=\textwidth]{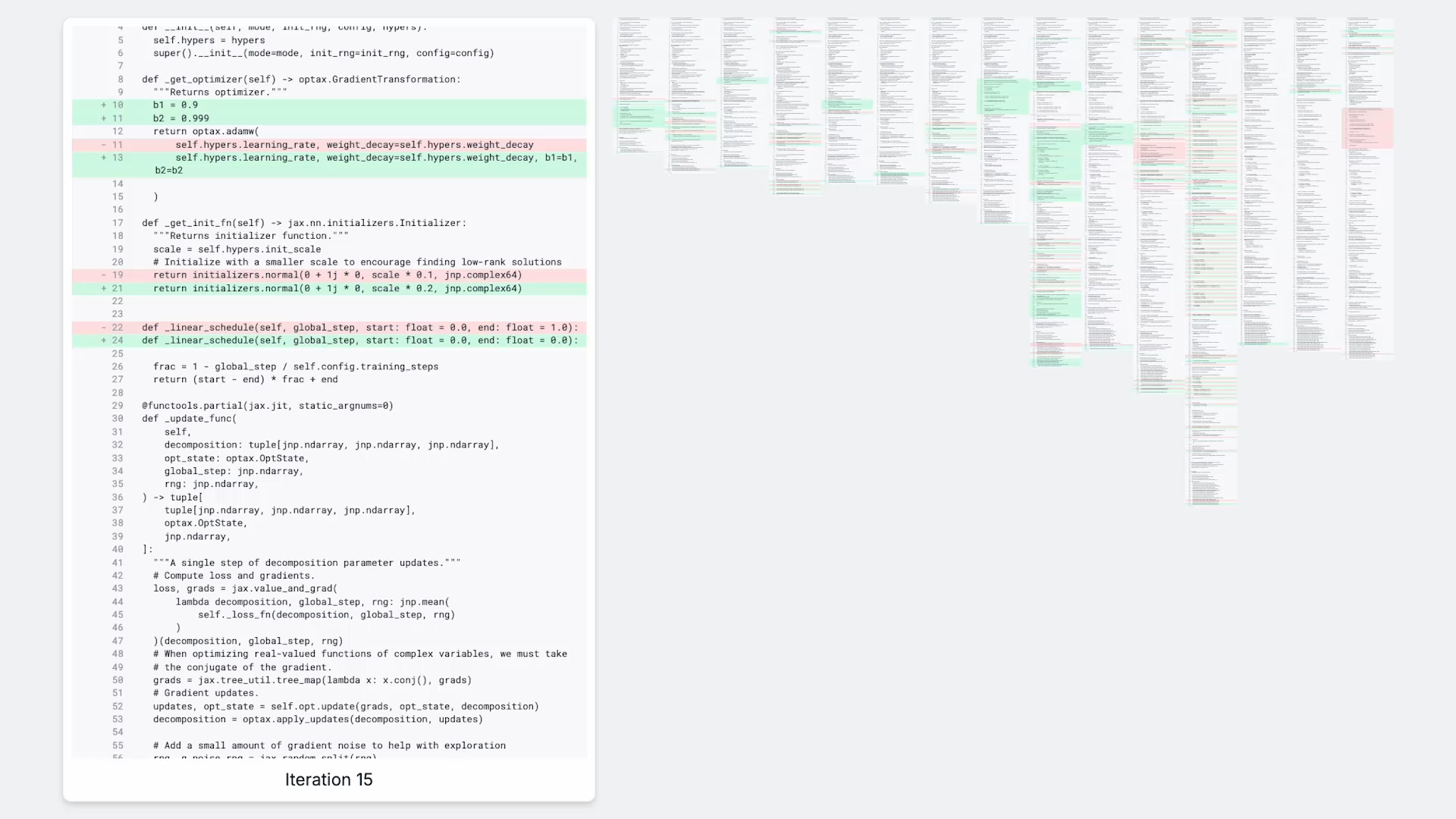}
\caption{A sequence of
 progressive changes proposed by AlphaEvolve to discover faster matrix multiplication algorithms. 
The smooth search space facilitates incremental improvement.}
\label{fig:AlphaEvolve}
\end{figure}

In programming, error recovery is comparatively tolerant and incremental. A~program may 
partially work (failing some tests but passing others), enabling informative feedback. Many 
mistakes are local, including modifying a line or two may preserve the overall behavior. Compilation 
provides fast sanity checking, and~multiple functionally equivalent implementations exist. 
These properties create a relatively smooth, navigable error landscape conducive to 
exploration, reward shaping, and~iterative refinement, as~illustrated in recent search-based 
systems such as~AlphaEvolve (see Figure~\ref{fig:AlphaEvolve}).

In contrast, error handling in formal proofs is brittle and discontinuous. A~single incorrect 
tactic can invalidate the entire proof state; newly generated subgoals may bear no relation to 
the intended argument; and there is no analog of partial correctness or fuzzy matching. A~
proof is either valid or not. Failures typically provide little gradient or structural signal: 
recovery is not local but may require a global reorganization of the proof strategy. 
Exploration is hazardous, as~small deviations often lead to unrecoverable~states.

As already observed in~\cite{AspertiSocial}, the~Curry--Howard correspondence---while 
theoretically elegant---does not eliminate these practical asymmetries. Proofs and programs 
play fundamentally different epistemic roles. Programs do things: they compute and 
produce observable outputs. Even a flawed program yields traces, errors, and~partial behavior 
that the model can exploit during training. Proofs, on~the contrary, are static objects whose 
purpose is simply to witness that a type (a proposition) is inhabited. A~``bugged'' proof 
rarely offers informative partial~feedback.

This conceptual divide also surfaces in the cautious attitude of the logic and ITP communities 
toward proofs-by-reflection~\cite{bertot2004,ssreflect}. Although~reflection is sound, its acceptance 
is bounded by concerns over opacity and the loss of human-inspectable reasoning structure. 
Such concerns underscore the broader point: the criteria by which we evaluate programs 
(efficiency, modularity, and behavior) and proofs (clarity, minimality, and logical \mbox{transparency) diverge.}

Taken together, these observations help explain why methods effective in programming do not 
transfer straightforwardly to proving. The~structure of the search space, the~nature of errors, 
the granularity of supervision, and~the epistemic roles of the resulting artifacts differ 
in ways that make formal proof generation substantially harder for current~LLMs.

\subsection{Are LLMs More Naturally Suited to Learning Traditional or Formalized Mathematics?} 
\label{sec:LLMs_formal_vs_informal}
At the current state-of-the-art, traditional, prose mathematical reasoning aligns much more naturally with the capabilities of large language models than formal reasoning. This is not merely a consequence of limited tool integration or lack of supervision in formal settings, but~reflects deeper representational and methodological differences. Traditional mathematics is embedded in narrative structure, expressed in natural language interleaved with symbolic notation, and~full of approximations, conventions, and~rhetorical shortcuts. This is exactly the kind of material LLMs are pretrained on: massive corpora that include textbooks, Wikipedia entries, StackExchange discussions, arXiv articles, and~lecture notes. In~such data, reasoning patterns appear repeatedly and are internalized through statistical association, even if they are not explicitly~formalized.

Formal mathematics, in~contrast, imposes a symbolic and syntactic discipline that makes reasoning brittle and opaque to models. Every formula must be perfectly typed; every proof step must obey inference rules; and nothing can be handwaved, while this resembles programming in its rigidity, code benefits from a smoother and more forgiving error landscape. When writing code, LLMs can rely on overlapping patterns, local feedback (e.g., compilation or tests), and~relatively high redundancy in syntax and semantics. In~formal mathematics, the~space of valid expressions is far narrower, and~even small deviations can render a proof completely invalid. For~a language model trained to predict the next token, this rigidity is difficult to internalize without external~feedback. 

The flexibility of the standard math prose compensates for this. In~a  typical argument, one can skip steps, use approximations, or~employ rhetorical cues like “clearly” or “it follows that” leaving part of the burden of reconstruction to the reader. This does not mean the reasoning is imprecise or vague; rather, it means that the structure of the reasoning is embedded in a smooth statistical surface that aligns well with how LLMs generalize from training. The~goal is not correctness by deduction, but~plausibility by association, and~in the traditional mathematical discourse, this often~suffices.

This is not to say that LLMs cannot be applied to formal mathematics. Indeed, some recent systems have demonstrated promising results by leveraging automatic binary feedback ,where a proof assistant can determine whether a generated proof successfully discharges the goal, while this form of supervision is coarse, namely it tells us only whether the whole proof is valid, it is reliable and scalable. Systems like DeepSeek-Prover, LeanDojo, and~Kimina-Prover make use of this mechanism, treating the proof assistant as an oracle. This enables techniques like rejection sampling, reinforcement learning from success/failure, and~large-scale self-play. However, the~lack of step-level feedback remains a bottleneck: models can be trained to generate plausible proof sketches, but~whether those steps align with a solvable tactic sequence is something only the prover can ultimately~verify.

%DISCUTERE AUOTFORMALIZATION COME ALTERNATIVA.
While the step-by-step interaction fits naturally with the architecture of interactive theorem provers, it should not be mistaken for a principled methodological choice. Rather, it reflects the current state-of-the-art and the practical constraints imposed by existing tools. There is no inherent reason why formal proofs could not be approached at a coarser level of granularity, especially if adopting a declarative proof style instead of a procedural one. We shall come back on this point in Section~\ref{sec:LLMs_formal_vs_informal}.

There is also another interesting issue worth discussing. A~promising but still largely unexplored direction (in the context of LLMs) is the adoption of declarative rather than procedural proof styles in the training and use of LLMs for formalized reasoning~\cite{isar_mizar}.
Procedural proofs rely on sequences of tactics, which are low-level, abstract, and~often opaque. Declarative proofs, by~contrast, more closely resemble natural mathematical argumentation: they articulate intermediate claims, proceed in a readable forward style, and~carry logical structure in the text itself~\cite{Asperti12pmc}. There is no hidden proof state, all information is externalized. This makes declarative proofs much easier to align with chain-of-thought prompting, allowing LLMs to express their reasoning explicitly in a form that mirrors human writing. It also reduces the abstraction gap between traditional and formalized reasoning, suggesting a promising hybrid mode in which models “think in math” not by mimicking tactics, but~by narrating their~understanding.

A crucial difficulty from this perspective is the so-called de Bruijn factor---a measure of the ``cost of formalization'' introduced by N.G. de Bruijn, the~creator of Automath~\cite{deBruijn1983}, one of the earliest formalized proof systems. It is defined as the ratio between the length of a formalized proof (measured in tokens, lines, or~size) and its human-written counterpart. This factor depends on several delicate aspects, including the complexity of the underlying mathematics and the maturity of the supporting proof libraries. Even in modern systems, the~de Bruijn factor is often estimated to be between 5 and 10~\cite{Asperti_usability,seventeen_provers}. This presents a serious challenge for generative approaches: not only do most language models operate within a fixed-size context window, but~generating long, syntactically constrained outputs while maintaining logical coherence becomes significantly more difficult as the length~increases.

\subsection{Do LLMs Possess a Notion of Computational State?}
\label{sec:LLMs_state}

In the previous sections, we examined coding, traditional mathematics, and~
formalized mathematics separately. We now address the third major question,
whether large language models possess---or can learn---a notion of 
computational state. Our analysis relies directly on the comparative 
observations from Section~\ref{sec:comparative_discussion}.

\subsubsection{What ``State'' Means Across~Domains}
Across the three domains considered, ``state'' plays very different roles.  
In coding, the~relevant state consists of execution traces, intermediate 
variables, scopes, and~program structure.  
In traditional mathematics, state is discursive and distributed: it is carried 
implicitly by the narrative, with~no external enforcement.  
In formalized mathematics, by~contrast, state is explicit, symbolic, and~
machine-maintained. A~proof assistant enforces a precise goal state and 
updates it after every~tactic.

From an architectural perspective, LLMs possess only a 
latent internal state (their hidden activations), but~no persistent 
working memory. Thus, three forms of state must be~distinguished:
\begin{enumerate}
    \item Internal latent state, encoded in the transformer 
    activations.
    \item External tool-maintained state, such as Lean's goal state 
    or a Python interpreter.
    \item Simulated state, represented linguistically by 
    chain-of-thought and other \\textual scaffolding.
\end{enumerate}

Let us also observe that transformers inherit from their RNN predecessors the idea of 
state-as-summarization. LSTMs maintained an explicit, persistent memory 
cell; transformers instead distribute state across the sequence via 
self-attention. This distributed representation is powerful, but~poorly 
matched to discrete symbolic transitions, a~point that becomes crucial in 
formalized~proving.

\subsubsection{Evidence from Coding~Systems}
In coding, LLMs exhibit a surprisingly robust ability to maintain a notion of ``program state'' but this ability is not grounded in a genuine internal simulation of execution. Rather, it emerges from the statistical regularities of code seen during pretraining. Programming languages are built on highly regular syntactic and structural patterns, in which variables are declared, updated, and~used in predictable ways; loops, functions, and~conditionals follow stable templates; and control flow is often expressed in left-to-right textual order. Transformers internalize these patterns as correlations in token sequences, and~during generation they exploit them to produce continuations that behave as if they were tracking a state~machine.

This pattern-based fluency is reinforced by the nature of coding supervision.
\mbox{Section~\ref{sec:proving_vs_coding}} shows that coding environments provide 
LLMs with rich external signals that function as state surrogates.  
Execution traces, unit tests, compiler diagnostics, and~other external signals expose models to partial, local, and~incrementally informative feedback. When a snippet of code fails one test but passes others, the~model receives a gradient pointing toward the failing component; the surrounding code may remain valid. This produces a smooth error landscape, where many small perturbations still yield runnable code, and~failure typically remains localized. Systems like AlphaCode and AlphaEvolve benefit enormously from this environment; search, RL, and~evolutionary strategies all thrive when the reward surface is dense and~incremental.

Thus, LLMs do not maintain the program state as a discrete symbolic object; instead, they reconstruct it implicitly through the manifold of textual patterns associated with the evolution of valid code. Even this ``soft'' notion of state is unexpectedly powerful, where it allows the model to preserve variable invariants, maintain consistent control flow, and~generate multi-step algorithmic structures without explicit memory. But~it remains fragile and context-dependent: when the syntactic or semantic structure deviates from familiar templates, the~illusion of state-tracking breaks~down.

\subsubsection{Evidence from Formalized~Mathematics}
Formal provers impose the strongest notion of state.  
Systems such as Kimina-Prover, Lyra, and~DeepSeek-Prover must match the 
exact proof state maintained by the kernel, with~no tolerance for deviation.  
Feedback is sparse and binary: a tactic either transforms the state correctly 
or it fails catastrophically.  
Exploration is brittle, and~the model receives almost no indication of how a 
failed state should be~repaired.

A further structural challenge concerns the orientation of reasoning.  
Human mathematical discourse is typically forward, in which  
statements accumulate into conclusions.  
Interactive theorem provers, however, work primarily backward;  
from a goal, they generate subgoals.  
As noted in the literature on proof styles 
(e.g., \cite{Wiedijk03,isar_mizar}), backward reasoning imposes 
a branching structure on the proof search.  
Transformers, which represent the computational state in a distributed and soft manner, 
have problems with such branching symbolic dependencies, and~traditionally 
struggle with backward inference steps~\cite{forward_backward24}.
Forward declarative reasoning is easier for LLMs to express, but~is not the 
native mode of most procedural proof assistants. 
This mismatch further widens the gap between the two forms of~reasoning.

\subsubsection{Evidence from Traditional Mathematical~Reasoning}
In traditional mathematical reasoning, models exhibit an intriguing combination of competence and fragility. On~the one hand, systems such as Minerva and DeepSeek-R1 routinely generate multi-step derivations that look mathematically meaningful. They chain algebraic transformations, recognize standard problem templates, introduce intermediate claims at appropriate points, and~deploy rhetorical cues that structure an argument in a recognizably human style. This suggests that LLMs internalize broad statistical patterns of mathematical discourse, not formal deduction, but~the narrative signatures of how typical solutions~unfold.

Crucially, however, this form of reasoning relies almost entirely on an implicit state. The~model does not maintain a symbolic representation of the evolving mathematical situation; it relies instead on statistical cues absorbed during pretraining. Certain turns of phrase tend to precede algebraic rearrangements; specific lemmas often follow particular templates; long-range logical dependencies are approximated by short local patterns. In~this respect, mathematical reasoning mirrors what is observed in code generation: coding models appear to “track state” not through internal simulation, but~by exploiting systematic regularities in syntax, layout, and~naming. They do not emulate abstract machine execution; instead, they reproduce the linguistic footprints of state transitions. Analogously, when producing a chain of thought, LLMs do not manipulate a structured proof state; they continue the textual form of what a plausible argument typically looks like at that~stage.

This gives rise to a sharp asymmetry. Locally, models can produce highly convincing steps; globally, they may lose coherence. Empirical evidence shows both faces of this~behavior as follows:
\begin{itemize}
\item Evidence of capability. Many solutions by Minerva or DeepSeek-R1 exhibit multi-layered reasoning---case-splits, inequality chains, decompositions of structure---that often align with competition-level heuristics. Recent empirical studies have begun to probe state-tracking mechanisms in controlled settings, including permutation-based state tracking~\cite{TrackState25} and finite-state/algorithmic tasks~\cite{FiniteState25}, as~well as long-horizon execution and trace-generation stress tests~\cite{In-ContextComplexity2025,StateTrackBench2025}.
\item Evidence of fragility. The~very same systems frequently contradict themselves, introduce unjustified claims, or~fall into circular reasoning. As~emphasized in~\cite{proof_or_bluff,math-failures2025,Unfaithful2023}, even when models obtain the correct numerical answer, the~accompanying arguments often fail to meet minimal standards of rigor: logical fallacies, misapplied lemmas, and~algebraic inconsistencies remain common.
\end{itemize}

The result is a mode of reasoning that is powerful but fundamentally approximate---guided by distributional cues rather than anchored to a persistent and verifiable state. This contrasts to formalized mathematics, where each step is validated against an explicit proof state maintained by the prover's~kernel.

\subsubsection{Interpretive~Conclusion}

Large language models do maintain a form of internal state, but~this state is 
distributed, soft, and~probabilistic, rather than discrete or symbolically 
verifiable. This stands in sharp contrast to the explicit, persistent, and~
authoritative notion of state required by formalized proof systems, where every 
intermediate step must satisfy strict syntactic and semantic constraints. The~
resulting mismatch helps explain the divergent behavior of LLMs across informal 
mathematics, formal proofs, and~programming~tasks.

The difficulty is not conceptual. Transformer architectures are, in~principle, 
capable of representing and updating structured information over time. Rather, 
the obstacles are practical and systemic. Current training pipelines rarely 
provide supervision that constrains intermediate semantic states; feedback from 
formal environments is typically sparse and binary; and many proof-oriented 
interfaces expose limited or brittle access to evolving proof states. Moreover, 
autoregressive generation places the burden of long-range consistency on implicit 
internal representations, without~mechanisms for persistent symbolic memory or 
authoritative state~correction.

A comparative reading of the domains surveyed in this paper highlights how these 
constraints interact with task structure. In~programming, the~execution 
environment maintains an external computational state that the model need not 
represent internally, but~can interrogate or validate against at discrete 
checkpoints through tests, error messages, and~traces. In~traditional mathematics, 
by contrast, no canonical external state exists without formalization; reasoning 
is embedded in a flexible narrative that tolerates redundancy, approximation, and~
local inconsistency. Formal mathematics occupies an intermediate but more demanding 
position: a precise external state is defined by the proof assistant, yet models 
interact with it through interfaces that are unforgiving and often provide limited 
guidance about admissible intermediate~transitions.

Taken together, these observations suggest that current LLMs succeed when the 
environment either externalizes state or relaxes the need to maintain it precisely, 
and struggle when neither condition holds. Their apparent ability to “reason” in 
informal settings reflects the permissiveness of the task distribution rather than 
a robust internalization of symbolic state. Conversely, failures in formal proof 
construction expose the limits of relying on implicit, text-based surrogates for 
state in environments that demand exactness. This perspective clarifies why formal 
mathematics imposes qualitatively different cognitive and computational demands, 
and motivates the broader questions addressed in Section~\ref{sec:major_questions}.

%\subsubsection{Interpretive conclusion}
%LLMs do maintain a form of internal state, but it is \emph{distributed, soft, and probabilistic}.  
%This contrasts sharply with the \emph{discrete, symbolic, and verifiable} state required for formalized proofs.  
%The difficulty is not conceptual: transformers could, in principle, learn symbolic state transitions.  
%The obstacles are practical:
%\begin{itemize}
%    \item no dense step-level supervision for proof states;
%    \item sparse binary signals from proof assistants;
%    \item brittle failure modes in tactic-based systems;
%    \item autoregressive architectures that lack persistent memory;
%    \item absence of external grounding comparable to test suites.
%\end{itemize}
%Coding offers structured external scaffolding; traditional mathematics offers flexible narrative scaffolding; formal mathematics offers neither.  As a result, current LLMs can simulate state when the environment provides redundant cues, but they cannot reliably maintain symbolic state where the environment is unforgiving.

\section{Iterative Proof, Search, and~Revision~Loops}
\label{sec:iterative_loops}

Due to the persistent limitations remarked in the previous section, a~growing body of work explores how large language models can improve their own outputs
through iterative refinement. In~contrast to single-pass generation, these approaches
introduce an explicit or implicit feedback loop in which a model evaluates, critiques, or~repairs its own reasoning, possibly through integration of RL with 
self-rewarding techniques~\cite{self_rewarding25}.

Such mechanisms are especially relevant to mathematical
reasoning, where proofs often require backtracking, local corrections, and~multi-step
planning. 

Two broad paradigms have~emerged:
\begin{enumerate}
    \item Multi-pass refinement loops, 
where the model repeatedly generates a draft,
    critiques it, and~attempts a corrected version.
    \item Intra-pass self-correction, where verification and repair are performed within a single autoregressive generation window.
\end{enumerate}

The following subsections survey these approaches and evaluate their potential for
mathematical and formalized~reasoning.

\subsection{Multi-Pass Search and Refinement~Loops}

Multi-pass frameworks such as Self-Refine, Reflexion, and~related approaches operate by
explicitly structuring generation into alternating~phases:
\begin{enumerate}
    \item Propose an initial solution;
    \item Detect errors or weaknesses (via self-critique or an external tool);
    \item Refine or regenerate the solution from scratch or in part.
\end{enumerate}

In mathematical reasoning, this pattern mirrors human problem solving, where
a mathematician drafts a proof, inspects unclear or incorrect steps, and~revises the
argument until a coherent solution emerges. LLMs, however, lack an internal notion of
logical correctness, so error detection must be guided by heuristics, consistency checks,
or external verification. For~traditional mathematics, where reasoning is expressed
narratively, these loops can substantially improve performance. Weak or contradictory
steps can often be repaired by rephrasing, reordering, or~restructuring the argument,
and local contradictions may be fixed without rewriting the entire~proof.

When external tools are available, the~loop becomes even more effective. In~coding tasks,
execution traces, compilation errors, or~failed test suites serve as highly informative
signals that help models converge toward correct solutions. Some systems for formal
mathematics adopt a similar strategy, using the proof assistant as a binary oracle.
A tactic either succeeds or fails, allowing the model to eliminate unproductive branches.
However, because~this feedback is sparse and non-local, many multi-pass loops stagnate:
the model repeatedly proposes variations in an unprovable tactic sequence without
meaningfully improving its~trajectory.

Thus, while multi-pass refinement is powerful in domains with dense or interpretable
feedback, its impact on formalized proving remains constrained by sparse rewards, brittle
state transitions, and~the absence of intermediate correctness~signals.

\subsection{Intra-Pass Self-Correction and~Verification}

Intra-pass methods attempt to embed revision inside a single generation process.
Models such as EditCoder and certain configurations of CriticGPT perform local
insertions, deletions, or~justifications during decoding, effectively interleaving
generation \mbox{with~verification.} 

Such approaches aim to give LLMs a form of “lightweight” verification mode, similar to a
mathematician who stops mid-proof to double-check a step before proceeding. Instruction
tuning can encourage the model to switch between generation and
verification mode, allowing spontaneous self-checks during long~derivations.
A few examples are reported in Table~\ref{tab:intra-pass models}.

\begin{table}[H]
\small
\caption{Models and methods for intra-pass~self-correction.}
%\centering
\begin{tabularx}{\textwidth}{cCc}
\toprule
\textbf{Model/Method} & \textbf{Mechanism} & \textbf{Domain} \\
\midrule
Self-Refine & Multi-pass generate--critique--revise loop & General reasoning \\
\midrule
Reflexion & Iterative self-analysis and targeted re-generation & Code, math, planning \\
\midrule
EditCoder & Inline insertion/deletion corrections \linebreak  during decoding & Code generation \\
\midrule
CriticGPT (some configs) & Interleaved critiques within a single pass & Dialogue, QA \\
\bottomrule
\end{tabularx}
\label{tab:intra-pass models}
\end{table}

However, these techniques come with significant caveats. As~documented in studies such as
~\cite{reflexion,self-refine}, models may exhibit performative self-correction, where
they introduce artificial mistakes solely to demonstrate their ability to fix them, or~inflate their reasoning with unnecessary critical commentary. Without~external grounding,
this behavior degrades clarity, encourages verbosity, and~does not reliably improve
correctness.

Moreover, self-correction remains shallow, a model can detect surface inconsistencies
(e.g., mismatched variable names or algebraic misalignments), but~lacks an internal
representation of “proof state” that would allow it to assess whether a step truly
advances the argument. In~mathematics, where correctness is global and brittle, the~lack
of persistent symbolic state makes intra-pass verification~unreliable.

\subsection{Challenges and Opportunities for~Mathematics}

For traditional mathematical reasoning, iterative refinement---whether multi-pass or
intra-pass---can yield substantial gains. Narrative proofs allow for loose constraints:
an error can often be mitigated by rephrasing or elaborating a subsequent step. The~self-corrective behavior need not be perfectly aligned with the underlying mathematical
state; plausibility often~suffices.

For formalized mathematics, the~situation is fundamentally different. Proof assistants enforce
rigid states with no partial credit and no tolerance for ambiguity. Multi-pass loops can
benefit from the prover's binary success/failure signal, but~intra-pass methods, which
rely on implicit verification, struggle to operate under symbolic rigidity. Without~dense
intermediate feedback or explicit state tracking, iterative loops often explore dead
branches repeatedly rather than converging on a valid~proof.

Nevertheless, the~integration of self-correction with external verification remains a promising direction. Hybrid pipelines, where LLMs propose structured revisions and provers validate them, could bridge the gap between narrative reasoning and symbolic
verification, offering a pathway toward more stable, grounded refinement~loops.

\section{Conclusions}
\label{sec:conclusion}
Throughout this paper, we have compared the behavior of large language models across three distinct mathematical regimes: traditional mathematical reasoning, formalized reasoning with proof assistants, and~program synthesis. As~discussed in Sections~\ref{sec:comparative_discussion} and~\ref{sec:LLMs_formal_vs_informal}, the~gap between these regimes does not reflect a principled limitation of neural architectures. Rather, it arises from the radically different learning signals, representational constraints, and~forms of feedback present in each~environment.

Traditional mathematics aligns naturally with the distributional assumptions implicit in large-scale pretraining. Its discourse is redundant, flexible, and~rhetorically structured; reasoning steps may be abbreviated, implicit, or~conventional, but~still intelligible to a statistically trained model. Local slips rarely destroy global plausibility, and~many derivations admit multiple solutions, allowing techniques such as chain-of-thought prompting, self-consistency, and~reward-model reranking to compensate for imperfect intermediate reasoning. This dense, forgiving landscape explains why models such as Minerva and DeepSeek-R1 perform so strongly in this regime: the structure of the task is well matched to the inductive biases of transformer~architectures.

Formalized mathematics presents the opposite extreme. Proof assistants maintain a single kernel-enforced proof state that must be updated with perfect syntactic and logical precision at every step. As~examined in Section~\ref{sec:comparative_discussion}, even small deviations from this state lead to immediate failure, and~feedback is binary, sparse, and~typically provided only after a complete proof attempt. Systems such as Kimina-Prover, Lyra, and~DeepSeek-Prover mitigate these difficulties through large-scale self-play, rejection sampling, or~RL based on success/failure signals, yet they remain constrained by the brittleness of the underlying search space. The~contrast with programming tasks is striking: code is syntactically rigid, but~its semantics are observable and decomposable through execution traces, test suites, and~partial correctness signals. This gives code LLMs a form of externalized state that is unavailable in theorem proving, enabling smoother optimization and more reliable recovery from~errors.

A unifying theme across these settings is the role of state. In~coding, the~state is grounded in execution; in traditional mathematics, it is spread through the narrative structure of the argument; in formalized proofs, it is explicitly managed by the prover's kernel. LLMs, maintained only a weak, distributed, and~probabilistic internal notion of state. Their chain-of-thought serves as a textual surrogate rather than a faithful computational record, and~they lack persistent symbolic memory. As~noted in Section~\ref{sec:major_questions}, nothing in principle prevents future models from learning more explicit state representations---indeed, recurrent architectures such as LSTMs already provided such mechanisms---but current training pipelines do not supply the step-level, semantically informative supervision required to develop~them.

Our investigation of autoformalization (Section~\ref{sec:autoformalization}) further highlights this tension. Translating mathematical statements and proofs into formal representations is not only difficult for present-day systems but also epistemically delicate. Autoformalized statements can be useful for interoperability between traditional and formalized mathematics; proof sketches can effectively guide automated provers; and declarative proof formats can narrow the abstraction gap between human reasoning and machine verification. However, complete end-to-end autoformalization remains highly unstable: the combinatorial explosion induced by the de~Bruijn factor, the~rigidity of formal grammars, and~the sensitivity to small syntactic deviations all conspire to create a brittle search environment. Hybrid pipelines that combine LLM-generated sketches with automated provers appear to offer the most promising path~forward.

Together, these observations support a balanced conclusion. LLMs are not inherently incapable of symbolic reasoning, nor are they naturally aligned with it. Their strengths emerge in environments with high redundancy, soft constraints, and~rich feedback; their weaknesses appear when faced with brittle symbolic interfaces, sparse rewards, and~long-range dependencies that require stable internal state. Traditional mathematics sits squarely in the former regime, formalized mathematics in the latter, and~coding occupies a middle ground where semantics are executable and feedback is comparatively~dense.

Bridging this gap will require progress on several fronts. Beyond~larger and better-structured datasets of formalized proofs, there is a need for evaluation methodologies and benchmarks that more sharply disentangle genuine reasoning from benchmark-specific heuristics. This includes benchmarks designed to minimize template reuse, controlled ablations that remove superficial scaffolding, and~tasks that require consistent manipulation of evolving intermediate results. In~particular, the~question of whether models maintain a meaningful notion of computational or deductive state, central to reasoning in both mathematics and programming, could itself be addressed through carefully designed benchmarks that probe state persistence and transformation across long reasoning trajectories. Alongside advances in declarative proof languages, tighter model--prover interfaces, training objectives that communicate intermediate semantic information, and~architectures capable of maintaining persistent symbolic state, such diagnostic evaluations may help clarify which limitations are contingent and which are structural. None of these obstacles appears fundamental. The~trajectory traced by recent advances in both coding and formal reasoning suggests that, with~the right combination of supervision, architecture, tooling, and~evaluation, systems that unify traditional and formal mathematics may eventually emerge---systems in which the fluidity of mathematical discourse and the precision of formal methods are no longer in tension but mutually~reinforcing. %MDPI: We removed Section ''Declarations'', please confirm.

\vspace{6pt}

\authorcontributions{Conceptualization, A.A. and A.N.; methodology, A.A. and C.S.C.; investigation, A.A., A.N. and C.S.C.; writing---original draft, A.A., A.N. and C.S.C.; supervision, A.A. All authors have read and agreed to the published version of the manuscript.}

\funding{Research partially supported by the Future AI Research (FAIR) project of the National Recovery and Resilience Plan (NRRP), Mission 4 Component 2 Investment 1.3 funded from the European Union--NextGenerationEU, and~by the Geometry of Algorithms (GoA) project funded by the French National Research Agency (ANR).}

\institutionalreview{Not applicable.}

\informedconsent{ot applicable.}

\dataavailability{No new data were created or analyzed in this study.}

\acknowledgments{Some of the content of this article has been presented in a talk delivered by the first author at the 
Workshop in honor of Georges Gonthier, held at Inria Paris, on~
{23 June 2025}. %MDPI: We revised the date format, please confirm.
We thank the organizers, E. Tassi and
A. Mahboubi, for~offering us this opportunity.}

\conflictsofinterest{The authors declare no conflicts of interest.}

\begin{adjustwidth}{-\extralength}{0cm}
%\centering %% If there is a figure in wide page, please release command \centering, for Table, ``\textwidth" should be ``\fulllength"
%\bibliography{reasoning,aesthetic,background}
\reftitle{References}

\PublishersNote{}
\end{adjustwidth}

\end{document}